**Title:** Automated brain extraction of multi-sequence MRI using artificial neural networks

**Short title**: ANN-based brain extraction with HD-BET

Fabian Isensee MSc[1*], Marianne Schell MD[2*], Irada Tursunova MD[3*], Gianluca Brugnara MD[2], David Bonekamp MD[3], Ulf Neuberger MD[2], Antje Wick MD[4], Heinz-Peter Schlemmer MD PhD[3], Sabine Heiland PhD[2], Wolfgang Wick MD[4,5], Martin Bendszus MD[2], Klaus H Maier-Hein PhD[1], Philipp Kickingereder MD[2]

*(1) Medical Image Computing, German Cancer Research Center (DKFZ), Heidelberg, Germany*
*(2) Department of Neuroradiology, University of Heidelberg Medical Center, Heidelberg, Germany*
*(3) Department of Radiology, DKFZ, Heidelberg, Germany*
*(4) Neurology Clinic, University of Heidelberg Medical Center, Heidelberg, Germany*
*(5) German Cancer Consortium (DKTK) in the German Cancer Research Center (DKFZ), Heidelberg, Germany*

* shared first authorship

**Corresponding Author and Address for Reprint Requests:**

Philipp Kickingereder, MD MBA

Department of Neuroradiology, University of Heidelberg

Im Neuenheimer Feld 400, 69120 Heidelberg, Germany

Email: philipp.kickingereder@med.uni-heidelberg.de

Phone: +49 (0) 6221 56 39069, Fax: +49 (0) 6221 56 4673

**Acknowledgments**: PK was supported by the Medical Faculty Heidelberg Postdoc-Program and the Else Kröner-Fresenius Foundation (Else-Kröner Memorial Scholarship).

**Abstract length**: 195 words; **Manuscript length**: 4258 words

**Figures**: 5; **Tables**: 4; **Data Supplement**: 2084 words

## Abstract

Brain extraction is a critical preprocessing step in the analysis of MRI neuroimaging studies and influences the accuracy of downstream analyses. The majority of brain extraction algorithms are, however, optimized for processing healthy brains and thus frequently fail in the presence of pathologically altered brain or when applied to heterogeneous MRI datasets. Here we introduce a new, rigorously validated algorithm (termed HD-BET) relying on artificial neural networks that aims to overcome these limitations. We demonstrate that HD-BET outperforms six popular, publicly available brain extraction algorithms in several large-scale neuroimaging datasets, including one from a prospective multicentric trial in neuro-oncology, yielding state-of-the-art performance with median improvements of +1.16 to +2.11 points for the DICE coefficient and -0.66 to -2.51 mm for the Hausdorff distance. Importantly, the HD-BET algorithm shows robust performance in the presence of pathology or treatment-induced tissue alterations, is applicable to a broad range of MRI sequence types and is not influenced by variations in MRI hardware and acquisition parameters encountered in both research and clinical practice. For broader accessibility our HD-BET prediction algorithm is made freely available (http://www.neuroAI-HD.org) and may become an essential component for robust, automated, high-throughput processing of MRI neuroimaging data.

## Key words

neuroimaging, brain extraction, skull stripping, artificial neural networks

**Introduction**

Brain extraction, which refers to the process of separating the brain from non-brain tissues in medical images is a preliminary but critical step in many neuroimaging studies conducted using magnetic resonance imaging (MRI). Consequently the accuracy of brain extraction may have an essential impact on the quality of the subsequent analyses such as image registration (Kleesiek, et al., 2016; Klein, et al., 2010; Woods, et al., 1993), segmentation of brain tumors or lesions (de Boer, et al., 2010; Menze, et al., 2015; Shattuck, et al., 2001; Wang, et al., 2010; Zhang, et al., 2001; Zhao, et al., 2010), measurement of global and regional brain volumes (e.g. in neurodegenerative diseases and multiple sclerosis (Frisoni, et al., 2010; Radue, et al., 2015)), estimation of cortical thickness (Haidar and Soul, 2006; MacDonald, et al., 2000), cortical surface reconstruction (Dale, et al., 1999; Tosun, et al., 2006) and for planning of neurosurgical interventions (Leote, et al., 2018).

Manual segmentation is currently considered the "gold-standard" for brain extraction (Smith, 2002; Souza, et al., 2018). However, this approach is not only very labor-intensive and time-consuming, but also shows a strong inter- and intraindividual variability (Kleesiek, et al., 2016; Smith, 2002; Souza, et al., 2018) that could ultimately bias the analysis and consequently hamper the reproducibility of clinical studies. To overcome these shortcomings several (semi-) automated brain extraction algorithms have been developed and optimized over the last years (Kalavathi and Prasath, 2016). Their generalizability is however limited in the presence of varying acquisition parameters or in the presence of abnormal pathological brain tissue, such as brain tumors. Without additional manual correction, poor brain extraction can introduce errors in downstream analysis (Beers, et al., 2018).

Artificial neural networks (ANN) have recently been successfully applied to a multitude of medical image segmentation tasks. In this context, several approaches based on ANN have been proposed to improve the accuracy of brain extraction. However, these ANN algorithms have focused on learning brain extraction from training datasets either containing a collection of normal (or apparently normal) brain MRI from public datasets (Dey and Hong, 2018; Sadegh Mohseni Salehi, et al., 2017), or from a limited number of (single institutional) brain MRI with pathologies (Beers, et al., 2018; Kleesiek, et al., 2016). Therefore, generalizability of these ANN algorithms to complex multicenter datasets may be limited on unseen data with varying MR hardware and acquisition parameters, pathologies or treatment-induced tissue alterations. Moreover, most approaches up until now focused on processing precontrast T1-weighted MRI sequences, since it provides a good contrast between different brain tissues and is frequently used as standard space for registration of further image sequences (Han, et al., 2018; Iglesias, et al., 2011; Lutkenhoff, et al., 2014). However they fall short when it comes to processing other types of MRI sequences, which would however be desirable from a clinical and trial perspective.

To overcome these limitations we utilize MRI data from a large multicenter clinical trial in neuro-oncology (EORTC-26101 (Wick, et al., 2017; Wick, et al., 2016)) to train and independently validate an ANN for brain extraction (subsequently referred to as *HD-BET*). Specifically, we aimed to develop an automated method that (a) performs robustly in the presence of pathological and treatment-induced tissue alterations, (b) is not influenced by variations in MRI hardware and acquisition parameters, and (c) is applicable to independently process various types of common anatomical MRI sequence.

## Methods

### Datasets

Four different datasets including the MRI data from a prospective randomized phase II and III trial in neuro-oncology (EORTC-26101) (Wick, et al., 2017; Wick, et al., 2016) and three independent public datasets (LPBA40, NFBS, CC-359) (Puccio, et al., 2016a; Shattuck, et al., 2008; Souza, et al., 2018) were used for the present study. The characteristics of the individual datasets were as follows:

*EORTC-26101*

The EORTC-26101 study was a prospective randomized phase II and III trial in patients with first progression of a glioblastoma after standard chemo-radiotherapy. Briefly, the phase II trial evaluated the optimal treatment sequence of bevacizumab and lomustine (four treatment arms with single agent vs. sequential vs. combination) (Wick, et al., 2016) whereas the subsequent phase III trial (two treatment arms) compared patients treated with lomustine alone with those receiving a combination of lomustine and bevacizumab (Wick, et al., 2017). Overall, the EORTC-26101 study included n=596 patients (n=159 from phase II and n=437 from phase III) with n=2593 individual MRI exams acquired at 37 institutions within Europe. The study was conducted in accordance with the Declaration of Helsinki and the protocol was approved by local ethics committees and patients provided written informed consent (EudraCT# 2010-023218-30 and NCT01290939). Full study design and outcomes have been published previously (Wick, et al., 2017; Wick, et al., 2016). MRI exams were acquired at baseline and every 6 weeks until week 24, afterwards every 3 months. For the present analysis we included T1-w, cT1-w, FLAIR and T2-w sequences (either acquired 3D and/or with axial orientation) and excluded those with heavy motion artifacts or corrupt data. These criteria were fulfilled by n=10005 individual sequences (including n=2401 T1-w,

n=2248 T2-w, n=2835 FLAIR and n=2521 cT1-w sequences from n=2401 exams and n=583 patients) which were included for the present analysis.

*Public datasets*

We used three public datasets for independent testing. Specifically, we collected and analyzed data from (a) the single-institutional LONI Probabilistic Brain Atlas (LPBA40) dataset of the Laboratory of Neuro Imaging (LONI) consisting of n=40 MRI scans from individual healthy human subjects (Shattuck, et al., 2008), (b) the single-institutional Nathan Kline Institute Enhanced Rockland Sample Neurofeedback Study (NFBS) dataset consisting of n=125 MRI scans from individual patients with a variety of clinical and subclinical psychiatric symptoms (Puccio, et al., 2016a), and (c) the Calgary-Campinas-359 (CC-359) dataset consisting of n=359 MRI scans from healthy adults (Souza, et al., 2018). For each subject, the repositories contains an anonymized (de-faced) T1-w MRI sequence and a manually-corrected ground-truth brain mask.

**Brain extraction using competing algorithms**

All MRI sequences from each of the datasets were preprocessed identically. First all images were reoriented to the standard (MNI) orientation (fslreorient2std, FMRIB software library, http://fsl.fmrib.ox.ac.uk/fsl/fslwiki/FSL), followed by the application of reference brain extraction algorithms. We compare HD-BET to six publicly available and frequently used brain extraction algorithms, namely BET (Smith, 2002), 3dSkullStrip (Cox, 1996), BSE (Shattuck and Leahy, 2002), ROBEX (Iglesias, et al., 2011), BEaST (Eskildsen, et al., 2012) and MONSTR (Roy, et al., 2017) (see **Supplementary Methods 1** for detailed description). As we intend HD-BET to be used out of the box, we also apply the reference methods as they are provided with no dataset-specific adaptations. For all competing brain extraction algorithms (except MONSTR) the maximum allowed processing time was set to 60 min (to keep processing within an acceptable time frame and execution of the brain extraction process was aborted if an algorithm exceeded this time limit for processing a single MRI sequence). Since BET, 3dSkullStrip, BSE, ROBEX and BEaST have primarily been developed for processing of T1-w sequences, we did not perform brain extraction with these algorithms on any other sequence type (i.e. cT1-w, FLAIR or T2-w) that were available in the EORTC-26101 test set. MONSTR is capable of also processing cT1-w, FLAIR and T2-w sequences and we therefore use it to perform brain extraction on all available sequences of the EORTC-26101 test set. In summary, this setup results in a comparison against six competing algorithms for brain extraction on T1-w sequences (EORTC-26101 test, LPBA40, NFBS, CC-359) and additional comparison against MONSTR on the remaining MRI sequences (T2-w, cT1-w, FLAIR) on EORTC-26101 test.

**Defining a ground-truth (reference) brain mask**

A ground-truth reference brain mask is required to evaluate the accuracy of brain extraction algorithms. Moreover, for the purpose of the present study with development of the HD-BET algorithm for automated brain extraction these masks are required to train the algorithm (i.e. to learn this specific task), as well as for subsequent evaluation of its accuracy. A ground-truth reference brain mask for the T1-w sequences was already provided within the three public datasets (LPBA40, NFBS, CC-359), whereas for the EORTC-26101 we generated a radiologist-annotated ground-truth reference brain mask for T1-w sequences as follows: The brain mask generated by BET algorithm was selected as a starting point. For each brain mask, visual inspection and corrections were performed using ITK-SNAP (by applying the different capabilities of this tool, including region-growing segmentation and manual corrections (www.itksnap.org (Yushkevich, et al., 2006))). The manual correction took on average about 15 minutes per brain mask. Given the amount of data, only one rater per ground-truth reference mask was used. Similar to the provided brain masks, we defined the following criteria: (1) including all cerebral and cerebellar gray and white matter as well as the brainstem, (2) including the cerebrospinal fluid in the ventricles and the cerebellar cistern and (3) excluding the chiasma. In a second step, to enable the use of the HD-BET algorithm independently of the input MRI sequence type (i.e. not limited to T1-w sequences) we transferred the ground-truth reference brain masks within the EORTC-26101 dataset from T1-w to the remaining anatomical sequences i.e. cT1-w, FLAIR and T2-w sequences. First, all sequences were spatially aligned to the respective T1-w sequence by rigid registration with 6-degrees of freedom (Greve and Fischl, 2009; Jenkinson and Smith, 2001), resulting in a transformation matrix for each of them. Next, the transformation matrix was inversely back transformed to the individual sequence space of the c T1-w, FLAIR and T2-w sequences and applied to the ground-truth reference brain mask (within the space of the T1-w sequence) using nearest neighbor interpolation. Thereby a ground-truth brain mask was generated for the remaining sequences (i.e. c T1-w, FLAIR and T2-w) within the individual sequence space. Finally, visual inspection was performed for all brain masks to exclude registration errors.

**Artificial neural network (ANN)**

The topology of the ANN underlying the HD-BET algorithm was inspired by the U-Net image segmentation architecture (Ronneberger, et al., 2015) and its 3D derivatives (Çiçek, et al., 2016; Kayalibay, et al., 2017; Milletari, et al., 2016) and has recently been shown to have excellent performance in brain tumor segmentation both in an international competition (Isensee, et al., 2018) as well as in the context of a large-scale multi-institutional study (Kickingereder, et al., 2019). **Supplementary Methods 2** contain an extended description of the architecture, as well as the training and evaluation procedure. Briefly, the EORTC-26101 dataset was divided into a training and test set using a random split of the dataset (~2:1 ratio) with the constraint that all patients from each of the 37 institution were either assigned to the training or test set (to limit the potential of overfitting the HD-BET algorithm). By applying this split, the EORTC-26101 training set included data from n=25 institutions (n=6586 individual MRI sequences from n=1568 exams, n=372 patients) whereas the EORTC-26101 test set included data from the remaining n=12 institutions (n=3419 individual MRI sequences from n=833 exams, n=211 patients). In this context it is important to emphasize that the EORTC-26101 test set was entirely independent from the training set, as it is comprised of acquisitions from different institutions (and thus different MRI scanners / field strengths, see Table 1 for the detailed information on the individual MRI sequences, scanner types, field strengths) that are disjunct from the institutions in the training set. All MRI sequences from the training set of the EORTC-26101 cohort (i.e. T1-w, cT1-w, FLAIR and T2-w) were used to train and validate the HD-BET algorithm (with 5-fold cross-validation). For independent large-scale testing and application of the HD-BET algorithm (done by using the five models from cross-validation as an ensemble), all MRI sequences from the test set of the EORTC-26101 cohort (i.e. T1-w, cT1-w, FLAIR and T2-w) as well as the T1-w sequences of the LPBA40, NFBS and CC-359 datasets were used. For both training and testing, the HD-BET algorithm was blinded to the type of MRI sequence used as input (i.e. T1-w, cT1-w, FLAIR or T2-w) which allowed to develop an algorithm that is capable to perform brain extraction irrespective of the type of anatomical MRI sequence.

**Evaluation metrics**

To evaluate the performance of the different brain extraction algorithms we compared the segmentation results of the different brain extraction methods with the ground-truth reference brain mask from each individual sequence. Among the numerous different metrics for measuring the similarity of two segmentation masks we calculated a volumetric measure, the Dice similarity coefficient (DICE, (Dice, 1945)) and a distance measure, the Hausdorff distance. The DICE coefficient is a standard metric for reporting the performance of segmentation and measures the extent of spatial overlap between two binary images, ground-truth (GT) and predicted brain mask (PM). It is defined as twice the size of the intersection between two masks normalized by the sum of their volumes.

$$\text{DICE} = \frac{2|GT \cap PM|}{|GT| + |PM|} * 100$$

Its values range between 0 (no overlap) and 100 (perfect agreement). However, volumetric measures can be insensitive to differences in edges, especially if this difference leads to an overall small volume effect relative to the total volume. Therefore we used the Hausdorff distance (Taha and Hanbury, 2015) to measure the maximal contour distance (mm) between the two masks.

$$\text{d}(x \rightarrow y) = \max(d_i^{x \rightarrow y}), i = 1..N_x$$

$$\text{Hausdorff distance}(GT, M) = \max(d(GT \rightarrow RM), d(RM \rightarrow GT))$$

The smaller the Hausdorff distance, the more similar the images. Here we took the 95[th] percentile of the Hausdorff distance, which is widely used for example in the evaluation of brain tumor segmentation (Menze, et al., 2015), as it allows to overcome the high sensitivity of the Hausdorff distance to outliers.

**Statistical analysis**

The Shapiro-Wilk test was performed to compare all evaluation metrics (DICE coefficient, Hausdorff distance) obtained from the T1-w sequences among the different brain extraction algorithms for normality. We report descriptive statistics (median, interquartile range (IQR)) for DICE coefficient and Hausdorff distance for all brain extraction algorithms in each of the datasets. To test the general differences of the different brain extraction algorithms in terms of their DICE coefficient and Hausdorff distance, we used a non-parametric Friedman or Skilling-Mack test. The latter was used in presence of missing data that would prevent a list-wise comparison (missing data resulted from those instances where the brain mask from one of the six competing brain extraction algorithms was not generated after exceeding the predefined time limit of 60 min for processing a single T1-w sequence, no time limit was used for MONSTR. For post-hoc comparisons, one-tailed Wilcoxon matched-pairs signed-rank tests were used to assess the performance of the HD-BET algorithm in comparison to the six competing brain extraction methods. The p-values from all post-hoc tests within each of the dataset were corrected for multiple comparison using the Bonferroni adjustment. The effect sizes of the post-hoc comparisons were interpreted using the Cohen classification (≥0.1 for small effects, ≥0.3 for medium effects and ≥0.5 for large effects(Cohen, 1988)).

For all other imaging sequences analyzed within the EORTC-26101 dataset (i.e. cT1-w, FLAIR and T2-w) we report descriptive statistics (median, IQR) for DICE coefficient and Hausdorff distance.

All statistical analyses were performed with R version 3.4.0 (R Foundation for Statistical Computing, Vienna, Austria)). P-values <0.05 were considered significant.

**Data Availability**

The MRI data from the EORTC-26101 trial that were used for training and independent large-scale testing of the HD-BET algorithm are not publicly available and restrictions apply to their use. The MRI data from the LPBA40, NFBS and CC-359 datasets are publically available and information on download is provided within the respective references cited in the Method section. For broader accessibility we provide a fully functional version of the presented HD-BET prediction algorithm for download via http://www.neuroAI-HD.org.

## Results

Within the EORTC-26101 training set (consisting of n=6586 individual MRI sequences with pre- and postcontrast T1-weighted (T1-w, cT1-w), FLAIR and T2-weighted (T2-w) sequences from 1568 MRI exams in 372 patients acquired across 25 institutions (**Table 1**)) the HD-BET algorithm acquired the relevant knowledge to generate a brain mask irrespective of the type of MRI sequence and in the presence of pathologies. Independent application and testing of the HD-BET algorithm in the EORTC-26101 test set (consisting of n=3419 individual MRI sequences from 833 exams in 211 patients acquired across 12 institutions (**Table 1**)) demonstrated similar performance with median DICE coefficients of 97.6 (IQR, 97.0-98.0) on T1-w, 96.9 (IQR, 96.1-97.4) on cT1-w, 96.4 (95.2-97.0) on FLAIR and 96.1 (IQR, 95.2-96.7) on T2-w sequences. Corresponding median Hausdorff distances ($95^{th}$ percentile) were 2.7 mm (IQR, 2.2-3.3 mm) on T1-w, 3.2 mm (IQR, 2.8-4.1 mm) on cT1-w, 4.2 mm (IQR, 3.4-5.0 mm) on FLAIR and 4.4 mm (IQR, 3.9-5.0 mm) on T2-w (**Figure 1 and Table 2**). Moreover, the performance was confirmed upon testing the HD-BET algorithm in three independent public datasets (LPBA40, NFBS, CC-359) which are specifically designed to evaluate the performance of brain extraction algorithms. In contrast to the EORTC-26101 dataset, application of the HD-BET algorithm in these public datasets was restricted to T1-w sequences since no other type of MRI sequence was provided. Specifically, we yielded median DICE coefficients of 97.5 (IQR, 97.4-97.7) for LPBA40, 98.2 (IQR, 98.0-98.4) for NFBS and 96.9 (IQR, 96.7-97.1) for the CC-359 datasets with corresponding median Hausdorff distances ($95^{th}$ percentile) of 2.9 mm (IQR, 2.5-3.0 mm), 2.8 mm (IQR, 2.4-2.8 mm) and 1.7 mm (IQR, 1.4-2.0 mm) again confirming both reproducibility and generalizability of the performance of our HD-BET algorithm (**Supplementary Table 1**).

Next, we compared the performance of our HD-BET algorithm with six competing brain extraction algorithms on each dataset (EORTC-26101 test set as well as the public LPBA40, NFBS and CC-359 datasets). For all competing brain extraction algorithms (except MONSTR), comparison was restricted to T1-w sequences since they have primarily been developed for processing of T1-w sequences and not optimized for independent processing of other sequence types (i.e. cT1-w, FLAIR or T2-w). MONSTR was applied to all available MRI sequences. We applied uniform non-parametric testing due to the evidence of non-normal data distribution for the majority of measurements ($p<0.05$ on Shapiro-Wilk test for 49/56 measurements – **Supplementary Table 2**). The obtained first-level statistics showed a significant difference between the investigated brain extraction methods for both evaluation metrics (DICE coefficient, Hausdorff distance) in each dataset ($p < 0.001$ for all comparisons – **Supplementary Table 3**).

Specifically, within the EORTC-26101 test set post-hoc Wilcoxon matched-pairs signed-rank test revealed significantly higher performance of our HD-BET algorithm (for both DICE coefficient and Hausdorff distance) as compared to each of the six competing brain extraction algorithms (Bonferroni-adjusted $p<0.001$ for all comparisons) maintaining a large effect size in 83% of the tests (10/12 comparisons) and medium effect size in the remaining 17% (2/12 comparisons) (**Figure 2-3 and Table 3**). Similarly, within the three public datasets post-hoc Wilcoxon matched-pairs signed-rank tests again demonstrated significantly higher performance of our HD-BET algorithm (for both DICE coefficient and Hausdorff distance) as compared to each of the six competing brain extraction algorithms (Bonferroni-adjusted $p<0.001$ for all but two comparisons: only the Hausdorff distance of the FSL-BET algorithm in the LPBA40 dataset and the MONSTR algorithm in the NFBS dataset were not significantly different from our HD-BET algorithm with an Bonferroni-adjusted p=0.221 and p=1). Moreover, 91% of the tests (31/34 comparisons) revealed a high effect size and 9% (3/31 comparisons) a medium effect (**Figure 2-3 and Table 3**). The improvement yielded with the HD-BET algorithm as compared to all competing algorithms within the different datasets ranged from +1.16 to +2.50 for DICE and -0.66 to -2.51 mm for the Hausdorff distance (95$^{th}$ percentile) and was most pronounced in the EORTC-26101 dataset (**Table 4**). **Figure 4** and 5 depict representative cases for the brain algorithms and sequences at different DICE values (5$^{th}$ percentile and median) from the EORTC-26101 test set and highlights the challenges associated with brain extraction in the presence of pathology and treatment-induced tissue alterations.

Average processing time for brain extraction of a single MRI sequence required 32 seconds of processing with the HD-BET algorithm (Nvidia Titan Xp GPU). In contrast, average processing time of a single T1-w sequence with one of the six competing public brain extraction algorithms ranged from 3 seconds to 34.6 minutes (specifically, averages were 3 seconds for BSE, 17 seconds for BET, 1.4 minutes for ROBEX, 4.0 minutes for 3dSkullstrip, 10.7 minutes for BEaST and 34.5 minutes for MONSTR) on a 8-core Intel Xeon E5-2640 v3 CPU.

For broader accessibility we provide a fully functional version of the presented HD-BET prediction algorithm for download via http://www.neuroAI-HD.org.

## Discussion

Here we present a method (HD-BET) that enables rapid, automated and robust brain extraction in the presence of pathology or treatment-induced tissue alterations, is applicable to a broad range of MRI sequence types and is not influenced by variations in MRI hardware and acquisition parameters encountered in both research and clinical practice. We demonstrate generalizability of the HD-BET algorithm on the EORTC-26101 test set with MRI sequences originating from 12 different institutions covering all major MRI vendors with a broad variety of scanner types and field strengths as well as within three independent public datasets. Importantly the EORTC test set is independent from the EORTC training set, since the institutions from which the imaging data originate differ. The HD-BET algorithm yields state-of-the-art performance in both the EORTC-26101 test set as well as three publicly available reference datasets (LPBA40, NFBS, CC-359). This finding reflects the limitations of many existing brain extraction algorithms which are usually not optimized for processing heterogeneous imaging data with pathological tissue alterations or varying hardware and acquisition parameters (Fennema-Notestine, et al., 2006) and consequently may introduce errors in downstream analysis of MRI neuroimaging data (Beers, et al., 2018). We addressed this within our study by training (and independent testing) the HD-BET algorithm with data from a large multicentric clinical trial in neuro-oncology which allowed to design a robust and broadly applicable brain extraction algorithm that enables high-throughput processing of neuroimaging data. Moreover, the improvement in the brain extraction performance yielded by the HD-BET algorithm was most pronounced in the EORTC-26101 dataset, again reflecting the limitations of the competing brain extraction algorithms when processing heterogeneous imaging data with abnormal pathologies or varying acquisition parameters.

The HD-BET algorithm is able to perform brain extraction on various types of common anatomical MRI sequence without prior knowledge of the sequence type. From a practical point of view this is of particular importance since imaging protocols (and the types of sequences acquired) may vary substantially. The majority of brain extraction algorithms are optimized to process T1-w MRI sequences (Han, et al., 2018; Iglesias, et al., 2011; Lutkenhoff, et al., 2014) and fall short during processing of other types of MRI sequences (e.g. T2-w, FLAIR or cT1-w images). We addressed this shortcoming and demonstrate that the HD-BET algorithm also performs well on cT1-w, FLAIR or T2-w MRI and closely replicates the performance observed for brain extraction on T1-w sequences. Our algorithm also outperformed MONSTR, which is explicitly designed to do brain extraction in the presence of pathologies and on other than T1-w MRI sequences in the EORTC-26101 test set as well as the public LPBA40 and CC-359 test sets.

The runtime of the HD-BET algorithm for processing a single MRI sequence is in the order of half a minute with modern hardware, including all pre- and postprocessing steps. More advanced hardware would allow to further improve processing time, although the existing setup already performed well in comparison to the runtime of the other competing brain extraction algorithms. For example, the 2$^{nd}$ best performing algorithm in the EORTC-26101 test set (MONSTR) required on average more than 30 minutes for processing of a single MRI sequence.

We acknowledge that although many different brain extraction algorithms have been proposed and published, we essentially focused on the most commonly used algorithms. Moreover a case-specific tuning of parameters from these brain extraction algorithms may have allowed to improve their performance to some extent (Iglesias, et al., 2011; Popescu, et al., 2012). This is particularly the case for BEaST, where a mismatch between source and target domain can result in a significant drop in performance ((Eskildsen, et al., 2012; Novosad, et al., 2018)). Dataset-specific adaptations are however not a practical approach, especially in the context of high-throughput processing. Moreover, we acknowledge that manually correcting brain masks in a single case can take hours (Puccio, et al., 2016b). Although our approach with generating a ground truth brain mask in a large-scale dataset was more focused on correcting major errors (e.g. around pathologies, resection cavities or due to varying hardware or acquisition parameters), even imperfect ground truth labels can lead to high quality deep-learning segmentation algorithms when using the UNET-architecture that was employed in our study (Heller, et al., 2018). Moreover the competitiveness of our approach was rendered by testing on the public datasets (NFBS, CC-359, and LPBA40) where we confirmed the performance of our HD-BET algorithm against an independent high-quality ground truth. In addition, future studies will need to evaluate the performance of our HD-BET algorithm in a broader range of diseases in neuroradiology since our evaluation was essentially limited to cases with brain tumors (EORTC-26101 dataset) or cases with only mild or no structural abnormalities (LPBA40, NFBS, CC-359 dataset). However, given the broad phenotypic appearance (and associated post-treatment alterations) of brain tumors which were used for training the algorithm we are confident that HD-BET is equally applicable to the broad disease spectrum encountered in neuroradiology.

In conclusion, the developed and rigorously validated HD-BET algorithm enables rapid, automated and robust brain extraction in the presence of pathology or treatment-induced tissue alterations, is applicable to a broad range of MRI sequence types and is not influenced by variations in MRI hardware and/or acquisition parameters encountered in both research and clinical practice. Taken together, HD-BET is made publicly available via http://www.neuroAI-HD.org and may become an essential component for robust, automated, high-throughput processing of MRI neuroimaging data.

# References

Beers, A., Brown, J., Chang, K., Hoebel, K., Gerstner, E., Rosen, B., Kalpathy-Cramer, J. (2018) DeepNeuro: an open-source deep learning toolbox for neuroimaging. ArXiv e-prints.

Çiçek, Ö., Abdulkadir, A., Lienkamp, S.S., Brox, T., Ronneberger, O. (3D U-Net: learning dense volumetric segmentation from sparse annotation). In; 2016. Springer. p 424-432.

Cohen, J. (1988) Statistical power analysis for the behavioral sciences. Hillsdale, N.J. L. Erlbaum Associates.

Cox, R.W. (1996) AFNI: software for analysis and visualization of functional magnetic resonance neuroimages. Computers and biomedical research, an international journal, 29:162-73.

Dale, A.M., Fischl, B., Sereno, M.I. (1999) Cortical surface-based analysis. I. Segmentation and surface reconstruction. NeuroImage, 9:179-94.

de Boer, R., Vrooman, H.A., Ikram, M.A., Vernooij, M.W., Breteler, M.M.B., van der Lugt, A., Niessen, W.J. (2010) Accuracy and reproducibility study of automatic MRI brain tissue segmentation methods. NeuroImage, 51:1047-1056.

Dey, R., Hong, Y. (2018) CompNet: Complementary Segmentation Network for Brain MRI Extraction. ArXiv e-prints.

Dice, L.R. (1945) Measures of the Amount of Ecologic Association Between Species. Ecology, 26:297-302.

Eskildsen, S.F., Coupe, P., Fonov, V., Manjon, J.V., Leung, K.K., Guizard, N., Wassef, S.N., Ostergaard, L.R., Collins, D.L. (2012) BEaST: brain extraction based on nonlocal segmentation technique. NeuroImage, 59:2362-73.

Fennema-Notestine, C., Ozyurt, I.B., Clark, C.P., Morris, S., Bischoff-Grethe, A., Bondi, M.W., Jernigan, T.L., Fischl, B., Segonne, F., Shattuck, D.W., Leahy,

R.M., Rex, D.E., Toga, A.W., Zou, K.H., Brown, G.G. (2006) Quantitative evaluation of automated skull-stripping methods applied to contemporary and legacy images: effects of diagnosis, bias correction, and slice location. Human brain mapping, 27:99-113.

Frisoni, G.B., Fox, N.C., Jack, C.R., Jr., Scheltens, P., Thompson, P.M. (2010) The clinical use of structural MRI in Alzheimer disease. Nat Rev Neurol, 6:67-77.

Greve, D.N., Fischl, B. (2009) Accurate and robust brain image alignment using boundary-based registration. NeuroImage, 48:63-72.

Haidar, H., Soul, J.S. (2006) Measurement of Cortical Thickness in 3D Brain MRI Data: Validation of the Laplacian Method. Journal of Neuroimaging, 16:146-153.

Han, X., Kwitt, R., Aylward, S., Bakas, S., Menze, B., Asturias, A., Vespa, P., Van Horn, J., Niethammer, M. (2018) Brain extraction from normal and pathological images: A joint PCA/Image-Reconstruction approach. NeuroImage, 176:431-445.

Heller, N., Dean, J., Papanikolopoulos, N. (2018) Imperfect Segmentation Labels: How Much Do They Matter? arXiv e-prints.

Iglesias, J.E., Liu, C.Y., Thompson, P.M., Tu, Z. (2011) Robust brain extraction across datasets and comparison with publicly available methods. IEEE transactions on medical imaging, 30:1617-34.

Isensee, F., Kickingereder, P., Wick, W., Bendszus, M., Maier-Hein, K.H. (2018) Brain Tumor Segmentation and Radiomics Survival Prediction: Contribution to the BRATS 2017 Challenge. arXiv e-prints.

Jenkinson, M., Smith, S. (2001) A global optimisation method for robust affine registration of brain images. Medical image analysis, 5:143-56.

Kalavathi, P., Prasath, V.B.S. (2016) Methods on Skull Stripping of MRI Head Scan Images—a Review. Journal of Digital Imaging, 29:365-379.

Kayalibay, B., Jensen, G., van der Smagt, P. (2017) CNN-based segmentation of medical imaging data. arXiv preprint arXiv:1701.03056.

Kickingereder, P., Isensee, F., Tursunova, I., Petersen, J., Neuberger, U., Bonekamp, D., Brugnara, G., Schell, M., Kessler, T., Foltyn, M., Harting, I., Sahm, F., Prager, M., Nowosielski, M., Wick, A., Nolden, M., Radbruch, A., Debus, J., Schlemmer, H.P., Heiland, S., Platten, M., von Deimling, A., van den Bent, M.J., Gorlia, T., Wick, W., Bendszus, M., Maier-Hein, K.H. (2019) Automated quantitative tumour response assessment of MRI in neuro-oncology with artificial neural networks: a multicentre, retrospective study. Lancet Oncol, 20:728-740.

Kleesiek, J., Urban, G., Hubert, A., Schwarz, D., Maier-Hein, K., Bendszus, M., Biller, A. (2016) Deep MRI brain extraction: A 3D convolutional neural network for skull stripping. NeuroImage, 129:460-469.

Klein, A., Ghosh, S.S., Avants, B., Yeo, B.T.T., Fischl, B., Ardekani, B., Gee, J.C., Mann, J.J., Parsey, R.V. (2010) Evaluation of volume-based and surface-based brain image registration methods. NeuroImage, 51:214-220.

Leote, J., Nunes, R.G., Cerqueira, L., Loução, R., Ferreira, H.A. (2018) Reconstruction of white matter fibre tracts using diffusion kurtosis tensor imaging at 1.5T: Pre-surgical planning in patients with gliomas. European Journal of Radiology Open, 5:20-23.

Lutkenhoff, E.S., Rosenberg, M., Chiang, J., Zhang, K., Pickard, J.D., Owen, A.M., Monti, M.M. (2014) Optimized Brain Extraction for Pathological Brains (optiBET). PLoS One, 9.

MacDonald, D., Kabani, N., Avis, D., Evans, A.C. (2000) Automated 3-D Extraction of Inner and Outer Surfaces of Cerebral Cortex from MRI. NeuroImage, 12:340-356.

Menze, B.H., Jakab, A., Bauer, S., Kalpathy-Cramer, J., Farahani, K., Kirby, J., Burren, Y., Porz, N., Slotboom, J., Wiest, R., Lanczi, L., Gerstner, E., Weber, M.A., Arbel, T., Avants, B.B., Ayache, N., Buendia, P., Collins, D.L., Cordier, N., Corso, J.J., Criminisi, A., Das, T., Delingette, H., Demiralp, C., Durst, C.R., Dojat, M., Doyle, S., Festa, J., Forbes, F., Geremia, E., Glocker, B., Golland, P., Guo, X., Hamamci, A., Iftekharuddin, K.M., Jena, R., John, N.M., Konukoglu, E., Lashkari, D., Mariz, J.A., Meier, R., Pereira, S., Precup, D., Price, S.J., Raviv, T.R., Reza, S.M., Ryan, M., Sarikaya, D., Schwartz, L., Shin, H.C., Shotton, J., Silva, C.A., Sousa, N., Subbanna, N.K., Szekely, G., Taylor, T.J., Thomas, O.M., Tustison, N.J., Unal, G., Vasseur, F., Wintermark, M., Ye, D.H., Zhao, L., Zhao, B., Zikic, D., Prastawa, M., Reyes, M., Van Leemput, K. (2015) The Multimodal Brain Tumor Image Segmentation Benchmark (BRATS). IEEE transactions on medical imaging, 34:1993-2024.

Milletari, F., Navab, N., Ahmadi, S.-A. (V-net: Fully convolutional neural networks for volumetric medical image segmentation). In; 2016. IEEE. p 565-571.

Novosad, P., Collins, D.L., Alzheimer's Disease Neuroimaging, I. (2018) An efficient and accurate method for robust inter-dataset brain extraction and comparisons with 9 other methods. Human brain mapping, 39:4241-4257.

Popescu, V., Battaglini, M., Hoogstrate, W.S., Verfaillie, S.C., Sluimer, I.C., van Schijndel, R.A., van Dijk, B.W., Cover, K.S., Knol, D.L., Jenkinson, M., Barkhof, F., de Stefano, N., Vrenken, H., Group, M.S. (2012) Optimizing parameter choice for FSL-Brain Extraction Tool (BET) on 3D T1 images in multiple sclerosis. NeuroImage, 61:1484-94.

Puccio, B., Pooley, J.P., Pellman, J.S., Taverna, E.C., Craddock, R.C. (2016a) The preprocessed connectomes project repository of manually corrected skull-stripped T1-weighted anatomical MRI data. GigaScience, 5:45.

Puccio, B., Pooley, J.P., Pellman, J.S., Taverna, E.C., Craddock, R.C. (2016b) The preprocessed connectomes project repository of manually corrected skull-stripped T1-weighted anatomical MRI data. GigaScience, 5:45.

Radue, E.W., Barkhof, F., Kappos, L., Sprenger, T., Haring, D.A., de Vera, A., von Rosenstiel, P., Bright, J.R., Francis, G., Cohen, J.A. (2015) Correlation between brain volume loss and clinical and MRI outcomes in multiple sclerosis. Neurology, 84:784-93.

Ronneberger, O., Fischer, P., Brox, T. (U-net: Convolutional networks for biomedical image segmentation). In; 2015. Springer. p 234-241.

Roy, S., Butman, J.A., Pham, D.L., Alzheimers Disease Neuroimaging, I. (2017) Robust skull stripping using multiple MR image contrasts insensitive to pathology. NeuroImage, 146:132-147.

Sadegh Mohseni Salehi, S., Erdogmus, D., Gholipour, A. (2017) Auto-context Convolutional Neural Network (Auto-Net) for Brain Extraction in Magnetic Resonance Imaging. ArXiv e-prints.

Shattuck, D.W., Leahy, R.M. (2002) BrainSuite: an automated cortical surface identification tool. Medical image analysis, 6:129-42.

Shattuck, D.W., Mirza, M., Adisetiyo, V., Hojatkashani, C., Salamon, G., Narr, K.L., Poldrack, R.A., Bilder, R.M., Toga, A.W. (2008) Construction of a 3D probabilistic atlas of human cortical structures. NeuroImage, 39:1064-80.

Shattuck, D.W., Sandor-Leahy, S.R., Schaper, K.A., Rottenberg, D.A., Leahy, R.M. (2001) Magnetic Resonance Image Tissue Classification Using a Partial Volume Model. NeuroImage, 13:856-876.

Smith, S.M. (2002) Fast robust automated brain extraction. Hum Brain Mapp, 17:143-55.

Souza, R., Lucena, O., Garrafa, J., Gobbi, D., Saluzzi, M., Appenzeller, S., Rittner, L., Frayne, R., Lotufo, R. (2018) An open, multi-vendor, multi-field-strength brain MR dataset and analysis of publicly available skull stripping methods agreement. NeuroImage, 170:482-494.

Taha, A.A., Hanbury, A. (2015) An efficient algorithm for calculating the exact Hausdorff distance. IEEE transactions on pattern analysis and machine intelligence, 37:2153-63.

Tosun, D., Rettmann, M.E., Naiman, D.Q., Resnick, S.M., Kraut, M.A., Prince, J.L. (2006) Cortical reconstruction using implicit surface evolution: Accuracy and precision analysis. NeuroImage, 29:838-852.

Wang, L., Chen, Y., Pan, X., Hong, X., Xia, D. (2010) Level set segmentation of brain magnetic resonance images based on local Gaussian distribution fitting energy. Journal of Neuroscience Methods, 188:316-325.

Wick, W., Gorlia, T., Bendszus, M., Taphoorn, M., Sahm, F., Harting, I., Brandes, A.A., Taal, W., Domont, J., Idbaih, A., Campone, M., Clement, P.M., Stupp, R., Fabbro, M., Le Rhun, E., Dubois, F., Weller, M., von Deimling, A., Golfinopoulos, V., Bromberg, J.C., Platten, M., Klein, M., van den Bent, M.J. (2017) Lomustine and Bevacizumab in Progressive Glioblastoma. The New England journal of medicine, 377:1954-1963.

Wick, W., Stupp, R., Gorlia, T., Bendszus, M., Sahm, F., Bromberg, J.E., Brandes, A.A., Vos, M.J., Domont, J., Idbaih, A., Frenel, J.-S., Clement, P.M., Fabbro, M., Rhun, E.L., Dubois, F., Musmeci, D., Platten, M., Golfinopoulos, V., Bent, M.J.V.D. (2016) Phase II part of EORTC study 26101: The sequence of bevacizumab and lomustine in patients with first recurrence of a glioblastoma. Journal of Clinical Oncology, 34:2019-2019.

Woods, R.P., Mazziotta, J.C., R. Cherry, Simon. (1993) MRI-PET Registration with Automated Algorithm. Journal of Computer Assisted Tomography, 17:536-546.

Yushkevich, P.A., Piven, J., Hazlett, H.C., Smith, R.G., Ho, S., Gee, J.C., Gerig, G. (2006) User-guided 3D active contour segmentation of anatomical structures: significantly improved efficiency and reliability. NeuroImage, 31:1116-28.

Zhang, Y., Brady, M., Smith, S. (2001) Segmentation of brain MR images through a hidden Markov random field model and the expectation-maximization algorithm. IEEE transactions on medical imaging, 20:45-57.

Zhao, L., Ruotsalainen, U., Hirvonen, J., Hietala, J., Tohka, J. (2010) Automatic cerebral and cerebellar hemisphere segmentation in 3D MRI: Adaptive disconnection algorithm. Medical Image Analysis, 14:360-372.

**Author contribution**

PK, MS, FI, IT designed the study; PK, MS, IT, GB, DB, UN performed quality control and preprocessing of the magnetic resonance imaging data from the EORTC-26101 dataset; IT, MS generated and PK visually inspected the ground-truth reference brain masks in the EORTC-26101 dataset; FI and PK applied the competing brain extraction algorithms to all datasets; FI performed development, training and application of the artificial neural network; FI calculated the evaluation metrics; MS and PK performed statistical analysis; PK, MS, FI interpreted the findings with essential input from all coauthors; PK, MS, FI, IT prepared the first draft of the manuscript; all authors critically revised the manuscript for important intellectual content; all authors approved the final version of the manuscript.

**Competing interests**

FI: none; MS: none; IT: none, GB: none; DB: Activities not related to the present article: received payment for lectures, including service on speakers' bureaus, from Profound Medical Inc; UN: none; AW: none; HPS: Activities not related to the present article: received payment from Curagita for consultancy and payment from Bayer and Curagita for lectures, including service on speakers' bureaus; SH: none; WW: Activities not related to the present article: received research grants from Apogenix, Boehringer Ingelheim, MSD, Pfizer, and Roche, as well as honoraria for lectures or advisory board participation or consulting from BMS, Celldex, MSD, and Roche; MB: Activities not related to the present article: received grant support from Siemens, Stryker, and Medtronic, consulting fees from Vascular Dynamics, Boehringer Ingelheim, and B. Braun, lecture fees from Teva, grant support and lecture fees from Novartis and Bayer, and grant support, consulting fees, and lecture fees from Codman Neuro and Guerbet; KHMH: none; PK: none

## Tables

**Table 1.** Characteristics of the datasets analyzed within the present study.

|  | EORTC-26101 | | LPBA40 | NFBS | CC-359 |
|---|---|---|---|---|---|
|  | Training set | Test set |  |  |  |
| Patients (n) | 372 | 211 | 40 | 125 | 359 |
| MRI exams (n) | 1568 | 833 | 40 | 125 | 359 |
| MRI exams per patient (median, IQR) | 4 (3-6) | 4 (3-6) | 1 | 1 | 1 |
| Institutes (n) | 25 | 12 | 1 | 1 | 2 |
| Patients per institute (median, IQR) | 7 (4-15) | 11 (3-20) | 1 | 1 | 60/299 |
| MRI Sequence (n) |  |  |  |  |  |
| T1-w | 1568 | 833 | 40 | 125 | 359 |
| cT1-w | 1623 | 898 | - | - | - |
| FLAIR | 1940 | 895 | - | - | - |
| T2-w | 1455 | 793 | - | - | - |
| MR vendors (n) |  |  |  |  |  |
| Siemens | 535 | 395 | - | 125 | 120 |
| Philips | 350 | 157 | - | - | 119 |
| General Electric | 640 | 267 | 40 | - | 120 |
| Toshiba | 12 | - | - | - | - |
| Unknown | 31 | 14 | - | - | - |
| MR field strength (n) |  |  |  |  |  |
| 1.0 Tesla | - | 9 | - | - | - |
| 1.5 Tesla | 631 | 78 | 40 | - | 179 |
| 3.0 Tesla | 216 | 317 | - | 125 | 180 |
| 1.5 or 3 Tesla | 619 | 415 | - | - | - |
| Unknown | 104 | 14 | - | - | - |

**Table 2.** Descriptive statistics on brain extraction performance (median and interquartile range (IQR) for DICE-coefficient and Hausdorff distance) in the EORTC test set for the different MRI sequences (T1-w, cT1-w, FLAIR, T2-w) and the corresponding statistics of the Wilcoxon matched-pairs signed-rank tests comparing the performance of HD-BET and MONSTR.

| MRI sequence type | DICE coefficient | | | | | | Hausdorff distance (95th percentile) | | | | | |
|---|---|---|---|---|---|---|---|---|---|---|---|---|
| | HD-BET | | MONSTR | | Statistics | | HD-BET | | MONSTR | | Statistics | |
| | median | IQR | media | IQR | abs(Z) | p | median | IRQ | media | IRQ | abs(Z) | p |
| T1-w | 97.6 | (97.0 - 98.0) | 95.4 | (94.0 - 96.1) | 30.62 | <.001 | 3.3 | (2.2 - 3.3) | 4.43 | (3.71 - 5.79) | 26.72 | <.001 |
| cT1-w | 96.9 | (96.1 - 97.4) | 94.6 | (93.2 – | 26.48 | <.001 | 3.9 | (2.8 - 4.1) | 5.48 | (4.36 - 6.96) | 26.92 | <.001 |
| FLAIR | 96.4 | (95.2 - 97.0) | 92.4 | (91.0 – | 32.16 | <.001 | 5.0 | (3.4 - 5.0) | 8.15 | (6.00 - 11.0) | 31.30 | <.001 |
| T2-w | 96.1 | (95.2 - 96.7) | 93.1 | (92.0 – | 30.64 | <.001 | 5.0 | (3.9 - 5.0) | 8.0 | (5.78 - 10.0) | 29.47 | <.001 |

**Table 3.** Wilcoxon matched-pairs signed-rank tests comparing the performance (DICE coefficient, Hausdorff distance) of our HD-BET algorithm with six competing brain extraction algorithms. For every test we reported the absolute value of the Z-statistics [abs(Z)], the Bonferroni-adjusted p-value and the effect size [r] (with r values >0.1 corresponding to a small effect, 0.3 to a medium effect and 0.5 to a large effect size, (Cohen, 1988)).

| Dataset | variable | BET | | | 3DSkullStrip | | | BSE | | | Robex | | | BEaST | | | MONSTR | | |
|---|---|---|---|---|---|---|---|---|---|---|---|---|---|---|---|---|---|---|---|
| | | abs(Z) | p | r | abs(Z) | p | r | abs(Z) | p | r | abs(Z) | p | r | abs(Z) | p | r | abs(Z) | p | r |
| EORTC-26101 test set | DICE | 24.31 | <.001 | .60 | 29.39 | <.001 | .72 | 27.69 | <.001 | .68 | 26.96 | <.001 | .48 | 3.89 | <.001 | .78 | 30.62 | <.001 | .75 |
| | Hausdorff* | 27.14 | <.001 | .66 | 27.88 | <.001 | .68 | 29.18 | <.001 | .72 | 25.69 | <.001 | .46 | 28.16 | <.001 | .71 | 26.72 | <.001 | .66 |
| LPBA40 | DICE | 3.95 | <.001 | .44 | 7.7 | <.001 | .86 | 7.7 | <.001 | .86 | 7.26 | <.001 | .81 | 7.33 | <.001 | .82 | 7.12 | <.001 | .81 |
| | Hausdorff* | 2.03 | .221 | - | 7.7 | <.001 | .86 | 7.7 | <.001 | .86 | 3.94 | <.001 | .44 | 3.69 | .001 | .41 | 4.73 | <.001 | .54 |
| NFBS | DICE | 13.67 | <.001 | .86 | 13.67 | <.001 | .86 | 12.5 | <.001 | .79 | 13.65 | <.001 | .86 | 11.22 | <.001 | .71 | 2.87 | 1 | - |
| | Hausdorff* | 13.68 | <.001 | .87 | 13.68 | <.001 | .87 | 11.08 | <.001 | .70 | 13.63 | <.001 | .86 | 12.79 | <.001 | .81 | 9.53 | <.001 | .61 |
| CC-359 | DICE | 22.72 | <.001 | .85 | 23.02 | <.001 | .86 | 21.69 | <.001 | .81 | 17.82 | <.001 | .67 | 21.05 | <.001 | .79 | 23.17 | <.001 | .87 |
| | Hausdorff* | 22.97 | <.001 | .86 | 23.05 | <.001 | .86 | 21.57 | <.001 | .80 | 21.77 | <.001 | .81 | 22.64 | <.001 | .84 | 23.20 | <.001 | .87 |

<u>Annotation</u>: * = using the 95$^{th}$ percentile of the Hausdorff distance (mm)

**Table 4.** Improvement of the performance for brain extraction with the HD-BET algorithm on T1-w sequences. The difference for each of the competing algorithms (as compared to HD-BET) was calculated on a case-by-case basis and summarized for all algorithms for each dataset by calculating the median and interquartile range (IQR). Positive values for the change in DICE coefficient (i.e. higher values with HD-BET), and negative values for the change in the Hausdorff distance (i.e. lower values with HD-BET) indicate better performance.

|  | **DICE coefficient** | | **Hausdorff distance*** | |
| --- | --- | --- | --- | --- |
|  | Median | IQR | Median | IQR |
| **EORTC-26101 test set** | +2.50 | +1.47 - +4.26 | -2.46 | -4.82 - -1.41 |
| **LPBA40** | +1.16 | +0.62 - +4.30 | -0.66 | -4.28 - -0.14 |
| **NFBS** | +1,67 | +0,67 - +3.85 | -1.91 | -3.39 - -0.92 |
| **CC-359** | +2.11 | +1.02 - +3.88 | -2.51 | -3.86 - -1.43 |

Annotation: * = using the 95$^{th}$ percentile of the Hausdorff distance (mm)

# Figures

**Figure 1**. DICE coefficient and Hausdorff distance (95[th] percentile) obtained from the individual sequences (pre- and postcontrast T1-weighted (T1-w, cT1-w), FLAIR and T2-w) with our HD-BET algorithm and for MONSTR in the EORTC-26101 test set using violin charts (and superimposed box plots). Obtained median DICE coefficients were >0.95 for all sequences. The performance of brain extraction on cT1-w, FLAIR or T2-w in terms of DICE coefficient (higher values indicate better performance) and Hausdorff distance (lower values indicate better performance) closely replicated the performance seen on T1-w (left column zoomed to the relevant range of DICE-values ≥0.9 and Hausdorff distance (HD95) ≤15 mm; right column depicting the full range of the data).

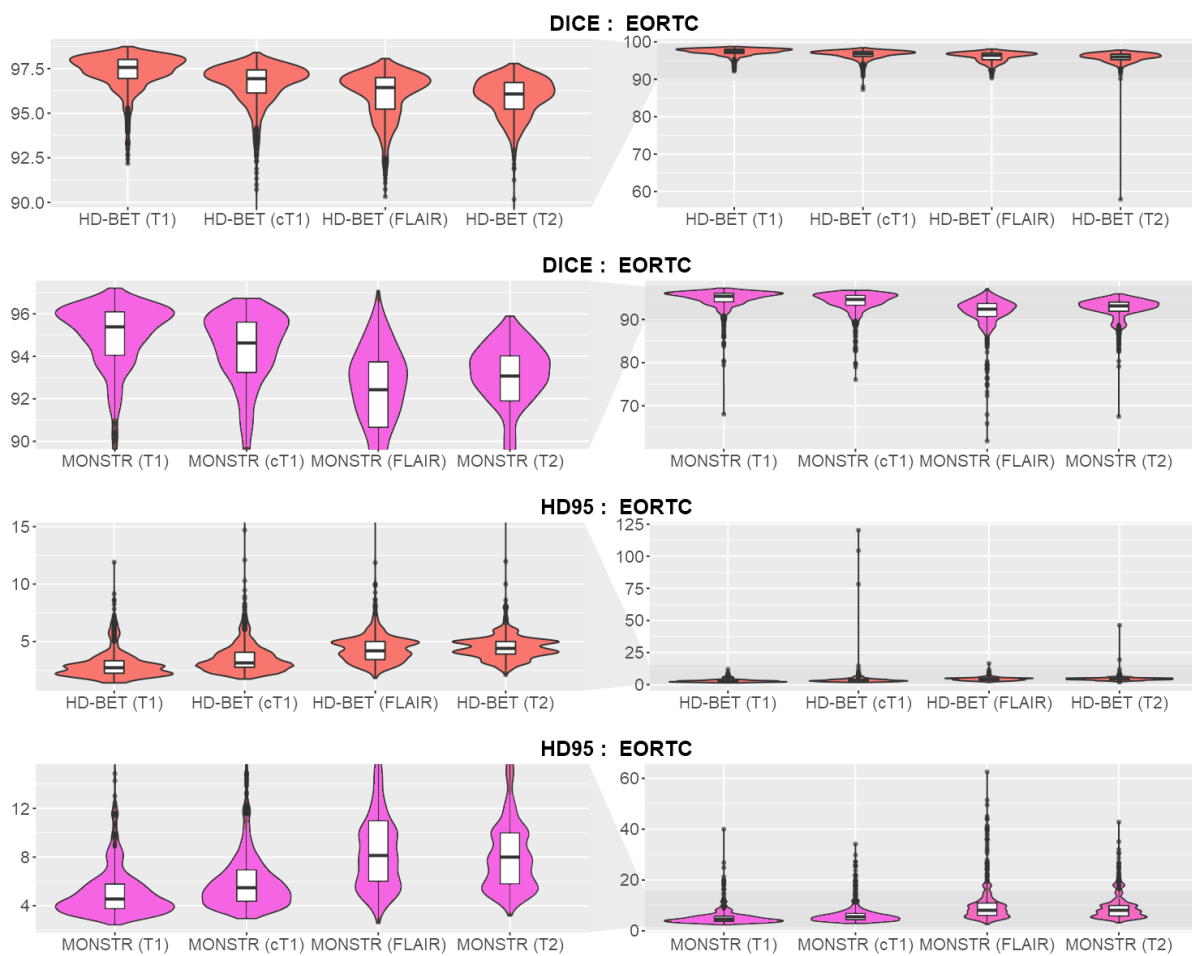

**Figure 2.** Comparison of DICE coefficients between our HD-BET brain extraction algorithm and the six public brain extraction methods for each of the test datasets using violin charts (and superimposed box plots) [higher values indicate better performance]. Obtained median DICE coefficients were highest for our HD-BET algorithm across all datasets (see left column visualizing the relevant range of DICE-values ≥0.9). Note the spread of the DICE coefficients, which is consistently lower for our HD-BET algorithm (right column visualizing the whole range of DICE values).

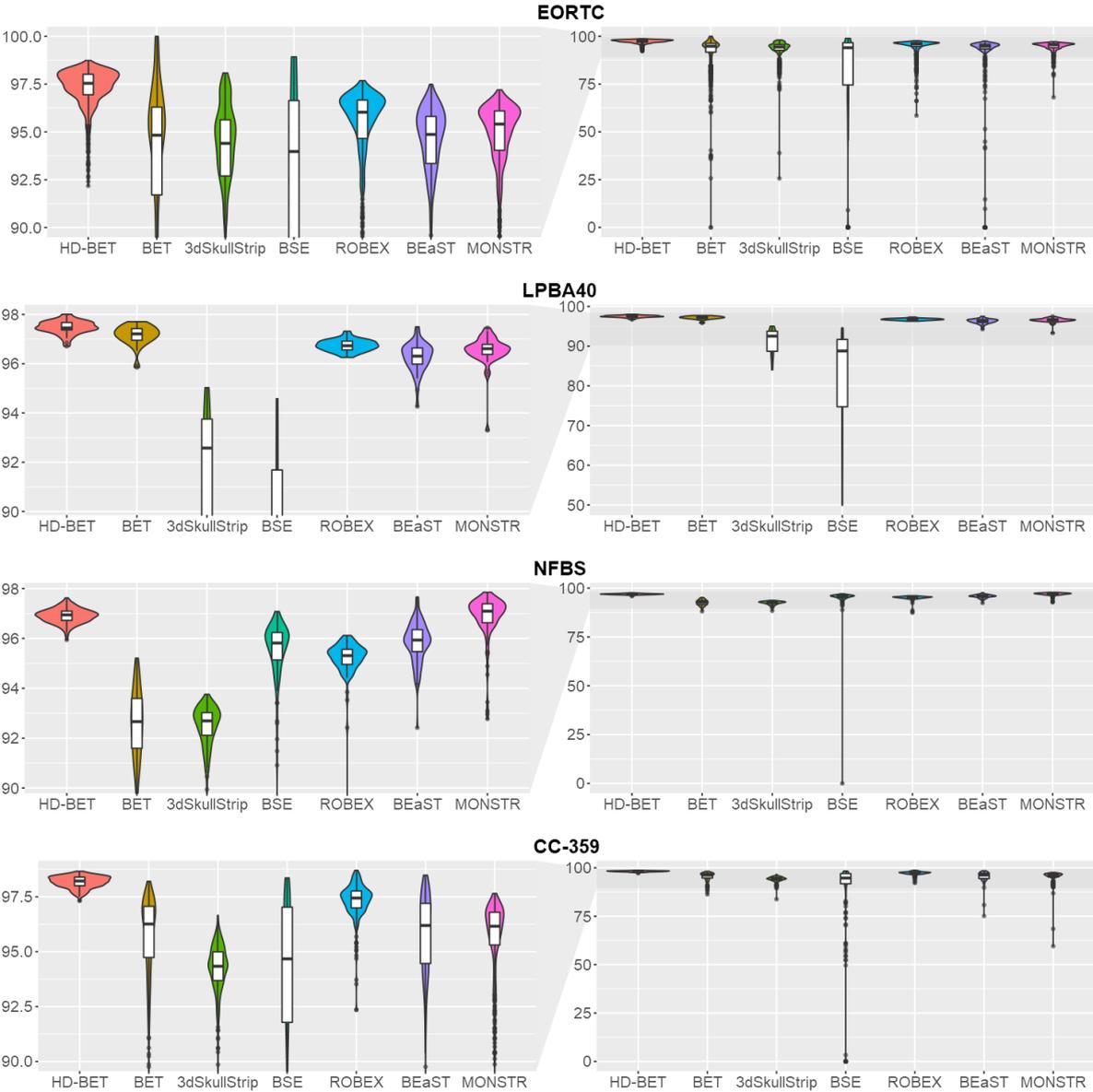

**Figure 3**. Comparison of Hausdorff distance (95[th] percentile) between our HD-BET algorithm and the six public brain extraction methods for each of the test datasets using violin charts (and superimposed box plots) [lower values indicate better performance]. The median Hausdorff distance was lowest for our HD-BET algorithm across all datasets (see left column visualizing the relevant range of Hausdorff distance ≤ 15 mm). Note the spread of the Hausdorff distance, which is consistently lower for our HD-BET algorithm (right column visualizing the whole range of values).

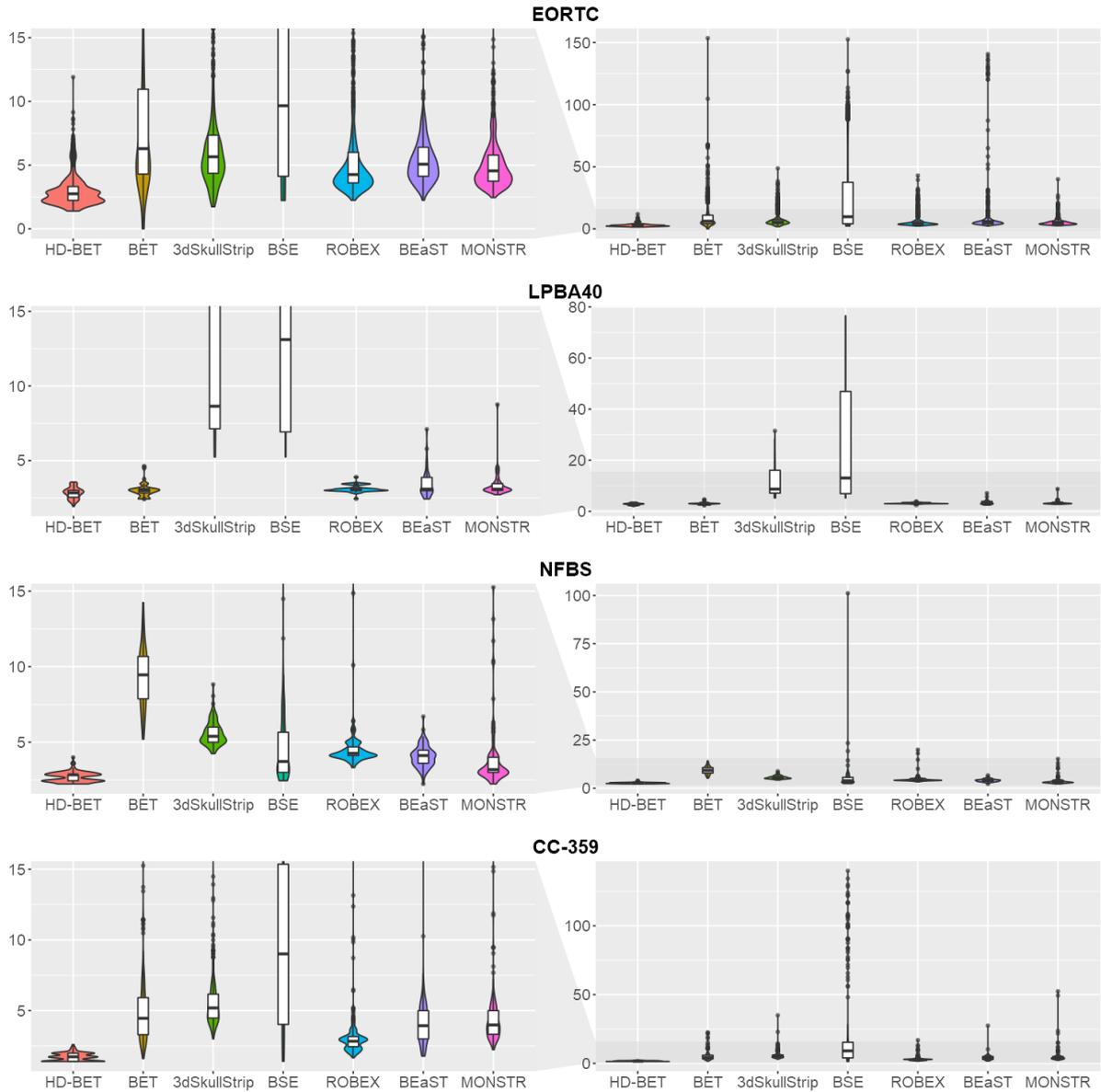

**Figure 4**. Representative cases showing the performance for T1-w images of the different brain extraction algorithms at the 5$^{th}$ percentile and the median DICE coefficients in the EORTC-26101 test set. Depicted in red the calculated brain masks from different brain extraction methods, in blue the ground-truth brain masks (for illustrative purposes only) and in pink their intersection. While BET, BEaST and MONSTR tend to underestimate the brain mask in these cases by removing brain tissue from the mask, 3DSkullStrip, BSE and ROBEX tend to overestimate by including non-brain tissue (e.g. skull, fat, nasal and orbital cavity) in the mask.

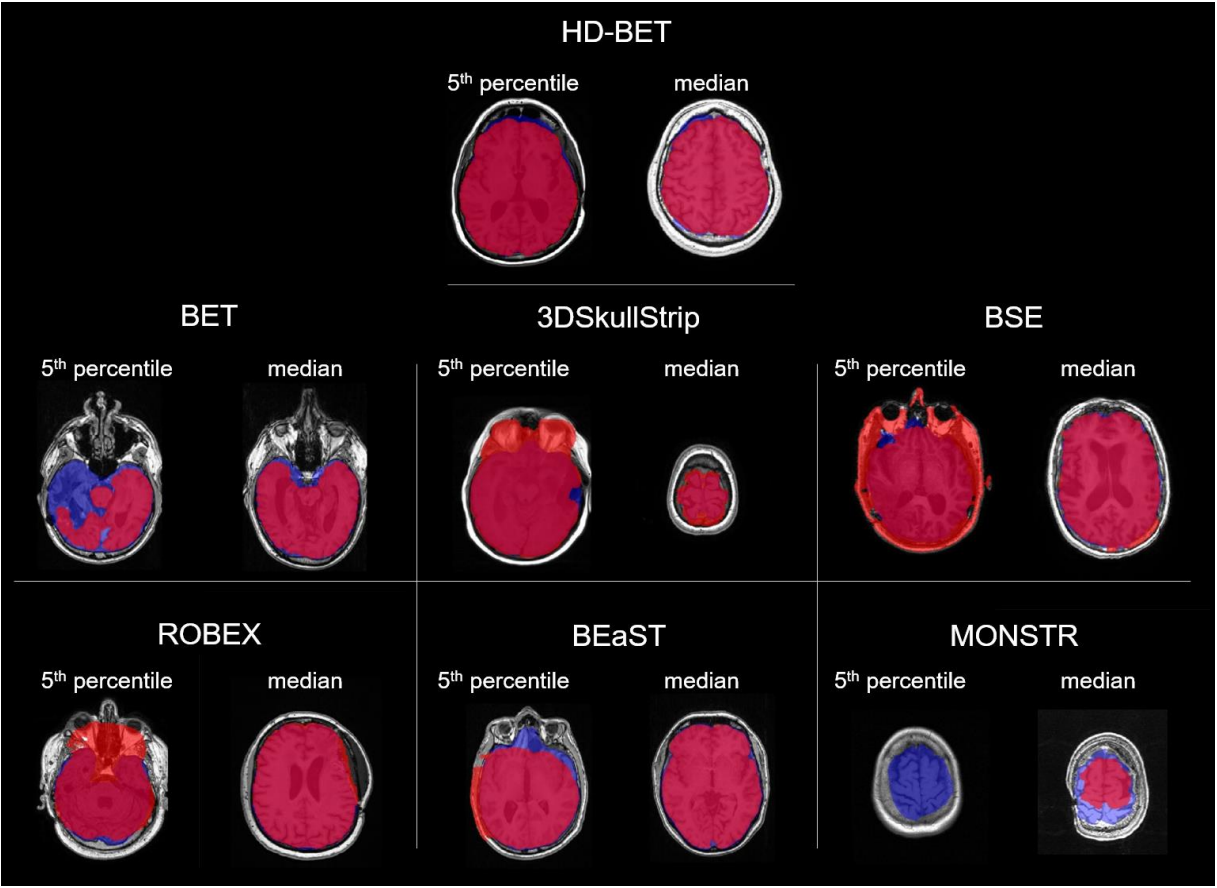

**Figure 5**. Representative cases showing the performances of HD-BET and MONSTR for cT1-w, FLAIR and T2-w images at 5th percentiles and medians of the DICE coefficients in the EORTC test set. Depicted in red the calculated brain masks (HD BET or MONSTR), in blue the ground-truth brain masks (for illustrative purposes only) and in pink their intersection. Similar to T1-w images MONSTR tends to underestimate in the brain mask in these cases by removing brain tissue from the masks and additionally for the 5th percentile in cT1-w and T2-w images tends to overestimate by including non-brain tissue around the nasal cavities.

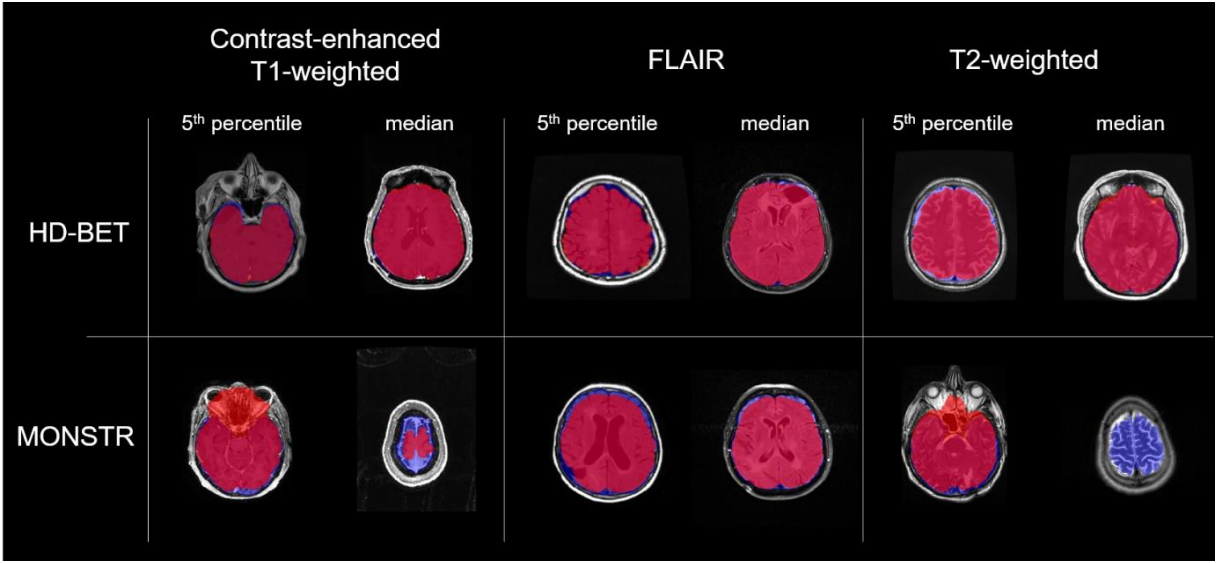

# Data supplement

# Automated brain extraction of multi-sequence MRI using artificial neural networks

*Table of contents:*

# A) Supplementary Methods

## 1. State-of-the-art brain extraction algorithms

Commonly available techniques such as the **Brain Extraction Tool** (**BET**(Jenkinson and Smith, 2001; Smith, 2002)) implemented in FSL (Jenkinson, et al., 2012; Woolrich, et al., 2009) and **3dSkullStrip** (part of the AFNI package (Cox, 1996)) are based on a deformable surface-based model (Dale, et al., 1999; Kelemen, et al., 1999) and create a brain mask though expanding and deforming of a defined template until its boundary fits into the surface of the brain (Kalavathi and Prasath, 2016; Smith, 2002; Souza, et al., 2018). Specifically, 3dSkullStrip is a modified version of BET and includes adjustments for avoiding the clipping of certain brain areas with two additional processing stages to ensure the convergence and reduction of the clipped area. Additionally is uses 3D edge detection.

**Brain Surface Extractor (BSE)** (Shattuck, et al., 2001) as part of the BrainSuite (Shattuck and Leahy, 2002) applies thresholding with morphology (Beare, et al., 2013; Hahn and Peitgen, 2000; Hohne and Hanson, 1992), in which the image is segmented by evaluating the intensity of the image pixels. In the next step, the uncertain voxels between brain and surrounding tissue are detected and subsequently eliminated through morphological filtering (Hohne and Hanson, 1992; Smith, 2002).

**ROBEX** (Iglesias, et al., 2011) is a method that uses affine registration of the image to a template to improve the performance. It combines a discriminative model that is trained to detect the brain boundaries and a generative model that ensures plausibility using a cost function.

Another example, named **BEaST** (Eskildsen, et al., 2012) is an atlas based method. It is built on nonlocal segmentation embedded in a multi-resolution framework and uses sum of squared differences to determine a suitable patch from a library of priors.

**MONSTR (Roy, et al., 2017)** uses non-local patch information from one or more atlases (where each atlas may contain multiple MRI sequences) to perform brain extraction. It was designed to be robust with respect to pathologies and is thus well suited for a comparison to HD-BET on the challenging EORTC-26101 test set. MONSTR can use combinations of MRI sequences as input or run on single MRI sequences independently. We apply MONSTR as provided by the authors (https://www.nitrc.org/projects/monstr). We use the TBIA atlas (version 1.1, 6 cases with T1-w, T2-w and FLAIR each) in all our experiments as this is the only provided atlas that covers all MRI modalities prevalent in the EORTC-26101 test set. To ensure a fair comparison to HD-BET, as well as to allow a consistent comparison of metrics, we process each modalities independently of each other (sequences in the EORTC test set are not registered). We use all patients from the TBIA atlas (n=6) and follow the authors recommendations for the remaining options.

All algorithms were used with standard parameters, except for BET where we added the options –R (for a more robust center estimation), -S and –B (to cleanup eye, optic nerve and neck voxels). Moreover all algorithms were applied as they are provided with no dataset-specific adaptations being made. This also includes the use of reference atlases. While this may result in suboptimal performance for some reference methods, this experimental setting is intentional. HD-BET is intended to be used out of the box and we therefore deem is a fair comparison to give the same treatment to the reference algorithms. Naturally, HD-BET was not re-trained on or otherwise optimized for the test set or any of the public datasets.

## 2. Artificial Neural Network (ANN)

All MRI sequences (and the corresponding brain masks) were downsampled to an isotropic spacing of 1.5x1.5x1.5 mm³ and normalized through z-scoring. The predicted output brain mask was linearly upsampled to the original resolution for evaluation.

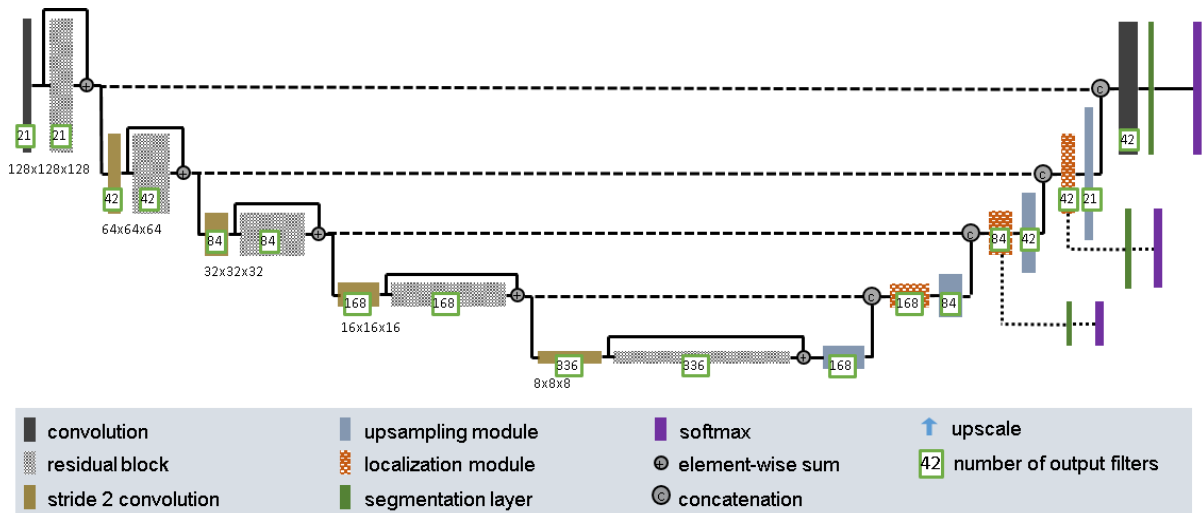

Network Architecture

The network architecture (depicted in Figure 1) shares similarities with our recent contribution (Isensee, et al., 2017) to the BraTS 2017 challenge (Menze, et al., 2015). It is inspired by the success of the U-Net architecture (Ronneberger, et al., 2015) and its 3D derivatives (Çiçek, et al., 2016; Kayalibay, et al., 2017; Milletari, et al., 2016).

U-Net sets itself apart from other segmentation networks (Havaei, et al., 2017; Kamnitsas, et al., 2017; Kleesiek, et al., 2016; Zhao, et al., 2018) by the use of an encoder and a decoder network that are interconnected with skip connections. Conceptually, the encoder network is used to aggregate semantic information at the cost of reduced spatial information. The decoder is the counterpart of the encoder that reconstructs the spatial information while being aware of the semantic information extracted from the encoder. Skip connections are used to transfer feature maps from the encoder to the decoder to allow for even more precise localization of the brain.

Heavy encoder, light decoder

Our instantiation of the U-Net utilizes pre-activation residual blocks (He, et al., 2016) in the encoder. Contrary to plain convolutions which learn a nonlinear transformation of the input, residual blocks learn a nonlinear residual that is added to the input. This allows the network by design to learn the identity function and ultimately allows the design of deeper architectures and improves the gradient flow. Here, a residual blocks consists of two 3x3x3 convolutional layers, each of which is preceded by instance normalization and a leaky ReLU nonlineariy.

We do not employ residual connections in the localization pathway. Here, each concatenation is followed by a 3x3x3 convolutional layer that is intended to recombine semantic and localization information, followed by a 1x1x1 convolution that halves the number of feature maps. We chose to upsample our feature maps by means of trilinear upsampling followed by a 3x3x3 convolution that again halves the number of feature maps. This approach allows us to leverage the benefits of convolutional upsampling (typically transposed convolution) without the risk of introducing checkerboard artifacts.

Large Input Patch Size

In order to maximize the amount of contextual information the encoder can aggregate, we train our network architecture with an input patch size of 128x128x128 voxels. At 1.5x1.5x1.5 mm³ voxel resolution this patch size almost covers an entire patient. Using such a large patch size enables the network to correctly reconstruct the brain mask even if large parts of the brain are missing due to a traumatic brain injury or the presence of a resection cavity.

Auxiliary Loss Layers

During training, the nature of gradient descent will optimize the network in a way that most quickly optimize the loss function. In the case of a U-Net like architecture such as the one presented here, this may lead to too simple decision making in the early stages of the training, i.e. solving most of the segmentation problem by forwarding local structures recognized early in the encoder to the decoder instead of making use of the entire receptive field the network can access. Additionally, gradients at the lower parts of the U shape are typically smaller due the nature of the chain rule. As a result, training the lower layers can be slow. We address both of these issues by integrating auxiliary loss layers deep into the network. These layers effectively create smaller versions of the desired segmentation, each of which are trained with its own loss layer and downsampled versions of the reference annotation.

Nonlinearity and Normalization

During model development we continuously observed dying ReLUs which motivated us to replace them with leaky ReLU nonlinearities throughout the network. Due to our small batch size, batch mean and standard deviation are unstable which may be problematic for batch normalization. For this reason we make use of instance normalization, which normalizes each sample in the batch independently of the others and which does not retain moving average estimates of batch mean and variance.

Training Procedure (with data augmentation)

The network architecture and hyperparameters were selected based on the results obtained from running a five-fold cross-validation on the training set of the EORTC-26101 cohort. Training was done with randomly sampled patches of size 128x128x128 voxels. These patches were cropped randomly from any of the four possible input modalities (T1, T2, FLAIR, cT1). The network is optimized using stochastic descent with the Adam algorithm (Kingma and Ba, 2014) (beta1=0.9, beta2=0.999, initial learning rate=1e-4) and a minibatch size of 2. The training took 200 epochs, where we define one epoch as the iteration over 200 training batches. An exponential learning rate decay was included to the training scheme by applying the following learning rate schedule: $\alpha_{epoch} = \alpha_0 * 0.99^{ep}$, where $\alpha_{epoch}$ represents the learning rate used at a specific epoch and $\alpha_0 = 10^{-4}$ is the initial learning rate.

Motivated by successful recent work (Drozdzal, et al., 2016; Isensee, et al., 2017; Kayalibay, et al., 2017; Milletari, et al., 2016; Sudre, et al., 2017) a soft dice loss formulation for training the network was used.

$$l_D(U,V) = -\frac{2}{|K|} \sum_{k \in K} \frac{\sum_i u_{i,k} v_{i,k}}{\sum_i u_{i,k} + \sum_i v_{i,k}}$$

Here, $u \in U$ denotes the voxels of the softmax output and $v \in V$ denotes a one hot encoding of the corresponding ground truth patch. Both $U$ and $V$ have shape shape Kx128x128x128 where $k \in K$ are the classes (background, brain). $i$ is used to index pixels in a patch (discarding spatial information; $i \in 128^3$).

As stated in the previous section, each auxiliary loss layer has its own dice loss term and is trained on a downsampled version of the reference annotation. The global loss is then computed as the weighted sum of these loss terms:

$$l = 0.25 l_{D,\frac{1}{4}} + 0.5 l_{D,\frac{1}{2}} + 1 l_{D,\frac{1}{1}},$$

where $l_{D,\frac{1}{4}}$ refers to the auxiliary loss layer that processes segmentations at $\frac{1}{4}$ resolution.

Data Augmentation

Due to their high capacity, neural networks tend to overfit given a limited amount of training data. Besides explicit regularization such as weight decay, stochastic gradient descent and dropout, implicit regularization in the form of data augmentation has proven to be very effective (Hernández-García and König, 2018). For this reason we apply a broad range of data augmentation techniques on the fly during training using a framework that was developed in our department and is available at http://github.com/MIC-DKFZ/batchgenerators). Hereby, U(a, b) denotes the uniform distribution on the interval [a, b].

- All input patches are mirrored randomly along all axes with probability 50%.
- 50% of patches are augmented with spatial transformations. These transformations include scaling, rotation and elastic deformation. Scaling is applied with a random scaling factor sampled from U(0.75, 1.25). Rotation is performed around all three axes with a random angle sampled from U(-180°, 180°) for each axis. Elastic deformation is implemented by sampling a grid of random, Gaussian distributed displacement vectors (μ=0, σ=1) which is then smoothed by a Gaussian smoothing filter with σ sampled uniformly from U(9, 13) and finally scaled by a randomly chosen scaling factor sampled uniformly from U(0, 900). We then apply the smoothed rescaled displacement vector field to the image and the corresponding segmentation via third order spline interpolation and nearest neighbor interpolation, respectively.
- Finally, we apply gamma augmentation to 50% of the patches. Gamma augmentation is done by transforming the voxel intensities to the interval [0, 1] and then applying the following equation for each voxel $I$.

$$I_{transformed} = I^\gamma$$

γ is hereby sampled from U(0.8, 1.5) once for each modality.
- 30% of the image patches are augmented with pixel-wise additive Gaussian Noise (μ=0, σ=0.2).
- A Gaussian blur filter with σ sampled from U(0.2, 1.5) was applied to 30% of the input patches.
- Since the gamma and Gaussian Noise augmentations alter the mean and standard deviation of the patches during training, whereas the network will only be presented z-score normalized inputs at test time, patches are renormalized to zero mean and unit variance before being fed into the network.

Postprocessing

Spurious misdetections are suppressed by means of connected component analysis: All segmented voxels that do not belong to the largest connected component are removed from the brain mask.

Evaluation

During evaluation, we apply data augmentation in the form of mirroring the data along all axes. Due to the fully convolutional nature of our network, we process entire images one at a time, alleviating the need for stitching patches together.

The prediction of the brain masks was performed in both training set (EORTC-26101 training set) and the four test sets (EORTC-26101 test set, LPBA40, NFBS and CC-359) using the following procedures. Predictions in the training set were generated from the samples within each of the holdout folds during 5-fold cross validation, whereas for test set patients we used the five networks obtained through the corresponding cross-validation as an ensemble to predict tumor segmentations. For the latter, softmax probabilities of the individual prediction of the five different networks are averaged to yield the final prediction. All computations performed using NVIDIA (NVIDIA Corporation, California, United States) Titan Xp graphics processing units.

**References**

Beare, R., Chen, J., Adamson, C.L., Silk, T., Thompson, D.K., Yang, J.Y.M., Anderson, V.A., Seal, M.L., Wood, A.G. (2013) Brain extraction using the watershed transform from markers. Frontiers in Neuroinformatics, 7:32.

Çiçek, Ö., Abdulkadir, A., Lienkamp, S.S., Brox, T., Ronneberger, O. (3D U-Net: learning dense volumetric segmentation from sparse annotation). In; 2016. Springer. p 424-432.

Cox, R.W. (1996) AFNI: software for analysis and visualization of functional magnetic resonance neuroimages. Computers and biomedical research, an international journal, 29:162-73.

Dale, A.M., Fischl, B., Sereno, M.I. (1999) Cortical surface-based analysis. I. Segmentation and surface reconstruction. NeuroImage, 9:179-94.

Drozdzal, M., Vorontsov, E., Chartrand, G., Kadoury, S., Pal, C. (2016) The importance of skip connections in biomedical image segmentation. Deep Learning and Data Labeling for Medical Applications: Springer. p 179-187.

Eskildsen, S.F., Coupé, P., Fonov, V., Manjón, J.V., Leung, K.K., Guizard, N., Wassef, S.N., Østergaard, L.R., Collins, D.L. (2012) BEaST: Brain extraction based on nonlocal segmentation technique. NeuroImage, 59:2362-2373.

Hahn, H.K., Peitgen, H.-O. (The Skull Stripping Problem in MRI Solved by a Single 3D Watershed Transform). In. Medical Image Computing and Computer-Assisted Intervention – MICCAI 2000; 2000; Berlin, Heidelberg. Springer Berlin Heidelberg. p 134-143.

Havaei, M., Davy, A., Warde-Farley, D., Biard, A., Courville, A., Bengio, Y., Pal, C., Jodoin, P.-M., Larochelle, H. (2017) Brain tumor segmentation with deep neural networks. Medical image analysis, 35:18-31.

He, K., Zhang, X., Ren, S., Sun, J. (Identity mappings in deep residual networks). In; 2016. Springer. p 630-645.

Hernández-García, A., König, P. (2018) Data augmentation instead of explicit regularization. arXiv preprint arXiv:1806.03852.

Hohne, K.H., Hanson, W.A. (1992) Interactive 3D segmentation of MRI and CT volumes using morphological operations. J Comput Assist Tomogr, 16:285-94.

Iglesias, J.E., Liu, C.Y., Thompson, P.M., Tu, Z. (2011) Robust brain extraction across datasets and comparison with publicly available methods. IEEE transactions on medical imaging, 30:1617-34.

Isensee, F., Kickingereder, P., Wick, W., Bendszus, M., Maier-Hein, K.H. (2017) Brain Tumor Segmentation and Radiomics Survival Prediction: Contribution to the BRATS 2017 Challenge. 2017 International MICCAI BraTS Challenge.

Jenkinson, M., Beckmann, C.F., Behrens, T.E.J., Woolrich, M.W., Smith, S.M. (2012) FSL. NeuroImage, 62:782-790.

Jenkinson, M., Smith, S. (2001) A global optimisation method for robust affine registration of brain images. Medical image analysis, 5:143-56.

Kalavathi, P., Prasath, V.B.S. (2016) Methods on Skull Stripping of MRI Head Scan Images—a Review. Journal of Digital Imaging, 29:365-379.

Kamnitsas, K., Ledig, C., Newcombe, V.F., Simpson, J.P., Kane, A.D., Menon, D.K., Rueckert, D., Glocker, B. (2017) Efficient multi-scale 3D CNN with fully connected CRF for accurate brain lesion segmentation. Medical image analysis, 36:61-78.

Kayalibay, B., Jensen, G., van der Smagt, P. (2017) CNN-based segmentation of medical imaging data. arXiv preprint arXiv:1701.03056.

Kelemen, A., Szekely, G., Gerig, G. (1999) Elastic model-based segmentation of 3-D neuroradiological data sets. IEEE transactions on medical imaging, 18:828-39.

Kingma, D.P., Ba, J. (2014) Adam: A method for stochastic optimization. arXiv preprint arXiv:1412.6980.

Kleesiek, J., Urban, G., Hubert, A., Schwarz, D., Maier-Hein, K., Bendszus, M., Biller, A. (2016) Deep MRI brain extraction: a 3D convolutional neural network for skull stripping. NeuroImage, 129:460-469.

Menze, B.H., Jakab, A., Bauer, S., Kalpathy-Cramer, J., Farahani, K., Kirby, J., Burren, Y., Porz, N., Slotboom, J., Wiest, R. (2015) The multimodal brain tumor image segmentation benchmark (BRATS). IEEE transactions on medical imaging, 34:1993-2024.

Milletari, F., Navab, N., Ahmadi, S.-A. (V-net: Fully convolutional neural networks for volumetric medical image segmentation). In; 2016. IEEE. p 565-571.

Ronneberger, O., Fischer, P., Brox, T. (U-net: Convolutional networks for biomedical image segmentation). In; 2015. Springer. p 234-241.

Roy, S., Butman, J.A., Pham, D.L., Alzheimers Disease Neuroimaging, I. (2017) Robust skull stripping using multiple MR image contrasts insensitive to pathology. NeuroImage, 146:132-147.

Shattuck, D.W., Leahy, R.M. (2002) BrainSuite: An automated cortical surface identification tool. Medical Image Analysis, 6:129-142.

Shattuck, D.W., Sandor-Leahy, S.R., Schaper, K.A., Rottenberg, D.A., Leahy, R.M. (2001) Magnetic Resonance Image Tissue Classification Using a Partial Volume Model. NeuroImage, 13:856-876.

Smith, S.M. (2002) Fast robust automated brain extraction. Hum Brain Mapp, 17:143-55.

Souza, R., Lucena, O., Garrafa, J., Gobbi, D., Saluzzi, M., Appenzeller, S., Rittner, L., Frayne, R., Lotufo, R. (2018) An open, multi-vendor, multi-field-strength brain MR dataset and analysis of publicly available skull stripping methods agreement. NeuroImage, 170:482-494.

Sudre, C.H., Li, W., Vercauteren, T., Ourselin, S., Cardoso, M.J. (2017) Generalised Dice overlap as a deep learning loss function for highly unbalanced segmentations. Deep Learning in Medical Image Analysis and Multimodal Learning for Clinical Decision Support: Springer. p 240-248.

Woolrich, M.W., Jbabdi, S., Patenaude, B., Chappell, M., Makni, S., Behrens, T., Beckmann, C., Jenkinson, M., Smith, S.M. (2009) Bayesian analysis of neuroimaging data in FSL. NeuroImage, 45:S173-86.

Zhao, G., Liu, F., Oler, J.A., Meyerand, M.E., Kalin, N.H., Birn, R.M. (2018) Bayesian convolutional neural network based MRI brain extraction on nonhuman primates. Neuroimage, 175:32-44.

# B) Supplementary Tables

**Supplementary Table 1.** Descriptive statistics (median, interquartile range (IQR) for the DICE-coefficient (upper panel; higher values indicate better performance) and Hausdorff distance (lower panel; lower values indicate better performance) of the different brain extraction algorithms on T1-w sequences across the different datasets.

| DICE coefficient | EORTC-26101 (test set) | | LPBA40 | | NFBS | | CC-359 | |
|---|---|---|---|---|---|---|---|---|
| Algorithm | median | IQR | median | IQR | median | IQR | median | IQR |
| HD-BET | 97.6 | (97.0 - 98.0) | 97.5 | (97.4 - 97.7) | 96.9 | (96.7 - 97.1) | 98.2 | (98.0 - 98.4) |
| BET | 94.8 | (91.7 - 96.3) | 97.2 | (97.0 - 97.4) | 92.7 | (91.6 - 93.6) | 96.3 | (94.7 - 97.1) |
| 3dSkullstrip | 94.4 | (92.7 - 95.6) | 92.6 | (88.7 - 93.8) | 92.7 | (92.1 - 93.0) | 94.4 | (93.7 - 95.0) |
| BSE | 94.0 | (74.6 - 96.6) | 88.9 | (74.7 - 91.7) | 95.8 | (95.1 - 96.2) | 94.7 | (91.8 - 97.0) |
| ROBEX | 96.0 | (94.7 - 96.7) | 96.7 | (96.6 - 96.9) | 95.3 | (95.0 - 95.6) | 97.4 | (97.0 - 97.8) |
| BEaST | 94.9 | (93.4 - 95.8) | 96.3 | (96.0 - 96.6) | 95.9 | (95.5 - 96.4) | 96.2 | (94.5 - 97.2) |
| MONSTR | 95.40 | (94.0 - 96.1) | 96.6 | (96.4 – 96.8) | 97.1 | (96,7 – 97.4) | 96.2 | (95.3 - 96.8) |

| Hausdorff distance | EORTC-26101 (test set) | | LPBA40 | | NFBS | | CC-359 | |
|---|---|---|---|---|---|---|---|---|
| Algorithm | median | IQR | median | IQR | median | IQR | median | IQR |
| HD-BET | 2.7 | (2.2 - 3.3) | 2.9 | (2.5 - 3.0) | 2.8 | (2.4 - 2.8) | 1.7 | (1.4 - 2.0) |
| BET | 6.3 | (4.3 - 11.0) | 3.0 | (2.8 - 3.1) | 9.4 | (7.9 - 10.7) | 4.4 | (3.3 - 5.9) |
| 3dSkullstrip | 5.7 | (4.4 - 7.3) | 8.7 | (7.1 - 16.1) | 5.4 | (5.0 - 6.0) | 5.2 | (4.5 - 6.2) |
| BSE | 9.7 | (4.1 - 37.5) | 13.1 | (6.9 - 46.9) | 3.7 | (3.0 - 5.7) | 9.0 | (4.0 - 15.4) |
| ROBEX | 4.2 | (3.6 - 6.0) | 3.0 | (3.0 - 3.2) | 4.2 | (4.1 - 4.7) | 2.8 | (2.4 - 3.2) |
| BEaST | 5.1 | (4.1 - 6.4) | 3.1 | (3.0 - 3.9) | 4.1 | (3.6 - 4.5) | 3.9 | (3.0 - 5.0) |
| MONSTR | 4.5 | (3.7 – 5.8) | 3.1 | (3.0-3.4) | 3.2 | (3.0 – 3.8) | 4.0 | (3.3 - 5.0) |

**Supplementary Table 2.** Shapiro-Wilk test for normality of DICE coefficient (upper panel) and Hausdorff distance (lower panel) for the different brain masks. Underlined p-values showed non-significant results.

| DICE coefficient | HD-BET | | | BET | | | 3DSkullStrip | | | BSE | | | Robex | | | BEaST | | | MONSTR | | |
|---|---|---|---|---|---|---|---|---|---|---|---|---|---|---|---|---|---|---|---|---|---|
| | Stats | df | p | Stats | df | p | Stats | df | p | Stats | df | p | Stats | df | p | Stats | df | p | Stats | df | p |
| EORTC-26101 test | .84 | 833 | <.001 | .57 | 833 | <.001 | .61 | 833 | <.001 | .66 | 833 | <.001 | .56 | 833 | <.001 | .27 | 792 | <.001 | .68 | 833 | <.001 |
| LPBA40 | .97 | 40 | .251 | .89 | 40 | .001 | .87 | 40 | <.001 | .85 | 40 | <.001 | .99 | 40 | .913 | .95 | 40 | .053 | .72 | 40 | <.001 |
| NFBS | .99 | 125 | .699 | .98 | 125 | .125 | .90 | 125 | <.001 | .14 | 125 | <.001 | .49 | 125 | <.001 | .97 | 125 | .003 | .71 | 125 | <.001 |
| CC-359 | .96 | 359 | <.001 | .87 | 359 | <.001 | .85 | 349 | <.001 | .44 | 359 | <.001 | .84 | 359 | <.001 | .78 | 358 | <.001 | .47 | 359 | <.001 |

| Hausdorff distance | HD-BET | | | BET | | | 3DSkullStrip | | | BSE | | | Robex | | | BEaST | | | MONSTR | | |
|---|---|---|---|---|---|---|---|---|---|---|---|---|---|---|---|---|---|---|---|---|---|
| | Stats | df | p | Stats | df | p | Stats | df | p | Stats | df | p | Stats | df | p | Stats | df | p | Stats | df | p |
| EORTC-26101 test | .82 | 833 | <.001 | .62 | 833 | <.001 | .68 | 833 | <.001 | .73 | 819 | <.001 | .60 | 833 | <.001 | .25 | 792 | <.001 | .63 | 833 | <.001 |
| LPBA40 | .96 | 40 | .162 | .83 | 40 | <.001 | .80 | 40 | <.001 | .78 | 40 | <.001 | .86 | 40 | <.001 | .80 | 40 | <.001 | .50 | 40 | <.001 |
| NFBS | .88 | 125 | <.001 | .99 | 125 | .593 | .92 | 125 | <.001 | .23 | 125 | <.001 | .37 | 125 | <.001 | .98 | 125 | .020 | .57 | 125 | <.001 |
| CC-359 | .87 | 359 | <.001 | .74 | 359 | <.001 | .58 | 349 | <.001 | .56 | 358 | <.001 | .52 | 359 | <.001 | .66 | 358 | <.001 | .30 | 358 | <.001 |

**Supplementary Table 3.** Friedman and Skilling-Mack test statistics for evaluating the general difference in terms of DICE coefficient in Hausdorff distance on T1-w images across the different brain extraction methods. Friedman test was used for the NFBS and LPBA40 dataset, whereas for the EORTC-26101 and CC-359 dataset the Skilling-Mack test with a simulated p-value of 10000 replications was used to prevent list-wise exclusion.

| Dataset | DICE coefficient | | Hausdorff distance | |
|---|---|---|---|---|
| EORTC-26101 test set (Skillings-Mack Statistic) | 2594.375 | <0.001 | 2649.406 | <0.001 |
| LPBA40 (Friedman test. $\chi^2$) | 203.325 | <0.001 | 169.247 | <0.001 |
| NFBS (Friedman test. $\chi^2$) | 597.723 | <0.001 | 533.894 | <0.001 |
| CC-359 (Skillings-Mack Statistic) | 478.303 | <0.001 | 525.100 | <0.001 |

**Supplementary Table 4.** Descriptive statistics of HD-BET in the EORTC-26101 **training set** using a 5-fold cross-validation (median, interquartile range (IQR) for the DICE-coefficient and Hausdorff distance

| MRI sequence type | DICE coefficient | | Hausdorff distance (95th percentile) | |
|---|---|---|---|---|
| | median | IQR | median | IRQ |
| T1-w | 97.0 | (96.3 - 97.7) | 3.3 | (2.5 - 4.4) |
| cT1-w | 96.4 | (95.5 - 97.1) | 3.9 | (3.0 - 5.0) |
| FLAIR | 96.0 | (95.0 - 96.8) | 5.0 | (3.7 - 5.2) |
| T2-w | 95.5 | (94.4 - 96.4) | 5.0 | (4.7 - 5.4) |

# C) Supplementary Figure

**Supplement Figure 1**. DICE coefficient and Hausdorff distance (95$^{th}$ percentile) obtained from the individual sequences (pre- and postcontrast T1-weighted (T1-w, cT1-w), FLAIR and T2-w) with our HD-BET in the EORTC-26101 training set (five fold cross validation) using violin charts (and superimposed box plots). Obtained median DICE coefficients were >0.95 for all sequences. The performance of brain extraction on cT1-w, FLAIR or T2-w in terms of DICE coefficient (higher values indicate better performance) and Hausdorff distance (lower values indicate better performance) closely replicated the performance seen on T1-w (left column zoomed to the relevant range of DICE-values ≥0.9 and Hausdorff distance ≤15 mm; right column depicting the full range of the data).

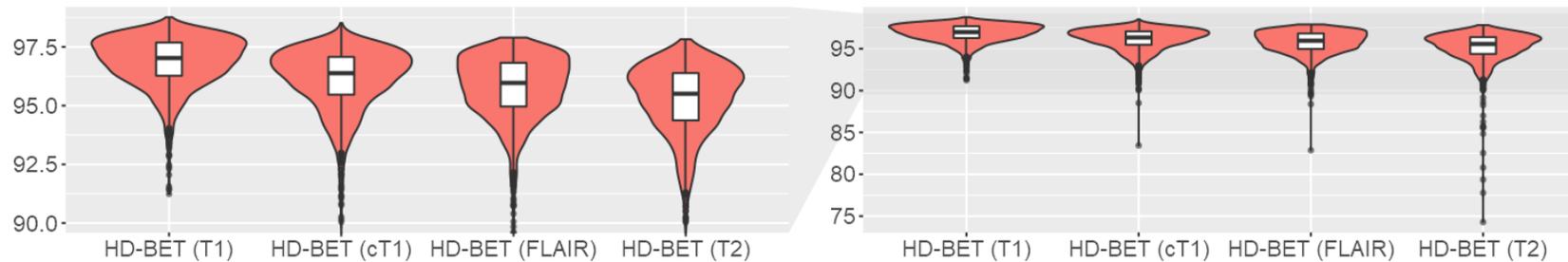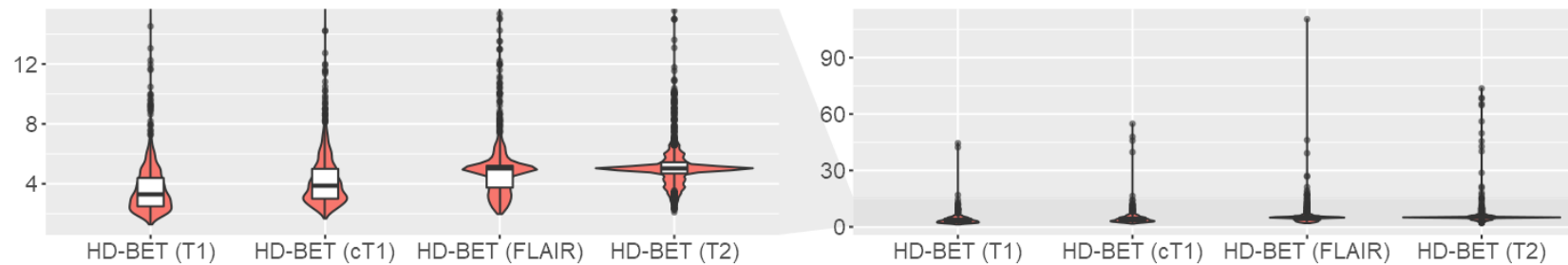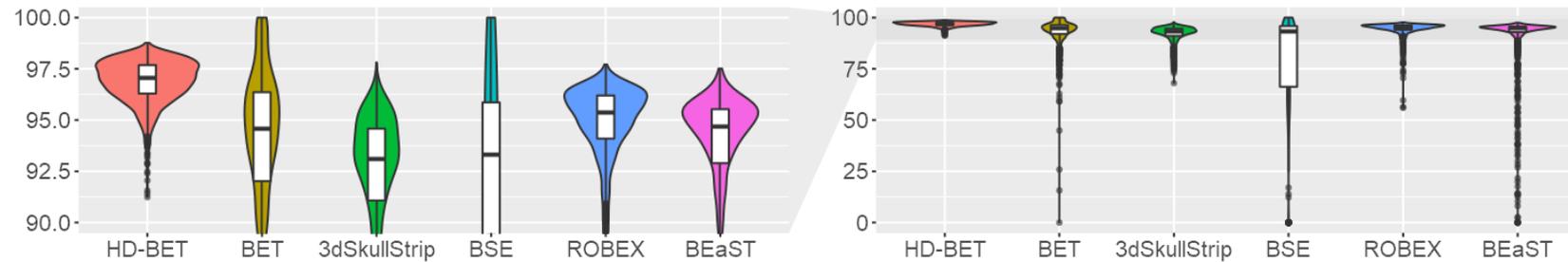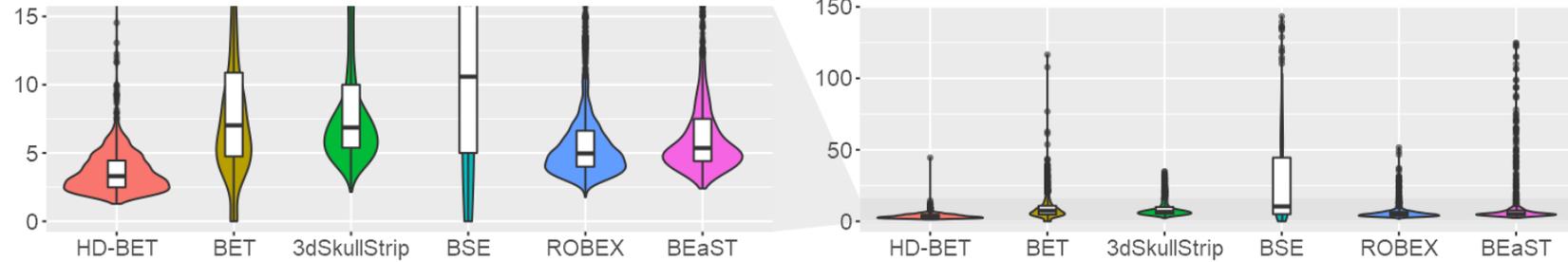